\documentclass[mnsc]{informs3aa} 

\usepackage[final]{showkeys}
\usepackage{subfigure}
\OneAndAHalfSpacedXI 
\usepackage{tabularx}
\usepackage{booktabs}

\newcolumntype{C}{>{\centering\arraybackslash}X}
\newcolumntype{P}[1]{>{\centering\arraybackslash}p{#1}}

\usepackage{endnotes}

\usepackage{longtable}
\usepackage{tabularx}
\usepackage{booktabs}
\usepackage{etex}
\usepackage{multirow}
\usepackage{array}
\usepackage{bbm}

 \usepackage{algpseudocode}

\newcommand{\op}{\mathrm{op}}

\newcommand{\mL}{\mathcal L}

\renewcommand{\hat}{\widehat}
\renewcommand{\tilde}{\widetilde}

\usepackage{amsbsy, amsfonts, amsgen, amsmath, amsopn, amssymb, amstext,
amsxtra, bezier, color, enumerate, graphicx, latexsym, verbatim,
pictexwd, supertabular, url, dsfont,leftidx, mathrsfs, appendix, setspace}
\usepackage{algorithm}
\usepackage{algorithmicx}
\usepackage{algpseudocode}
\makeatletter
\def\BState{\State\hskip-\ALG@thistlm}
\makeatother

\usepackage{natbib}
 \bibpunct[, ]{(}{)}{,}{a}{}{,}%
 \def\bibfont{\small}%
\usepackage[colorlinks=true,bookmarks=false,urlcolor=blue, citecolor=blue,linkcolor=blue,bookmarksopen=false,draft=false]{hyperref}

\usepackage{xcolor}
\usepackage{epstopdf}

\renewcommand{\hat}{\widehat}
\renewcommand{\tilde}{\widetilde}
\renewcommand{\bar}{\overline}

\definecolor{DSgray}{cmyk}{0,1,0,0}

\newcommand{\rwd}{\mathrm{Rwd}}
\newcommand{\reg}{\mathrm{Reg}}

\newcommand{\mH}{\mathcal H}

\newcommand{\mT}{\mathcal T}
		
\TheoremsNumberedThrough     
\ECRepeatTheorems

\EquationsNumberedThrough    

\MANUSCRIPTNO{} 

\begin{document}



\RUNTITLE{Optimal Network Revenue Management with Nonparametric Demand Learning}

\TITLE{Demand Balancing in Primal-Dual Optimization for Blind Network Revenue Management}

\ARTICLEAUTHORS{%
\AUTHOR{Sentao Miao\thanks{Author names listed in alphabetical order.}}
\AFF{Leeds School of Business, University of Colorado, Boulder, CO 80309, USA\\
sentao.miao@colorado.edu}
\AUTHOR{Yining Wang}
\AFF{Naveen Jindal School of Management, University of Texas at Dallas, Richardson, TX 75019, USA\\
yining.wang@utdallas.edu}
} 

\ABSTRACT{
{
This paper proposes a practically efficient algorithm with optimal theoretical regret which solves the classical network revenue management (NRM) problem with unknown, nonparametric demand. Over a time horizon of length $T$, in each time period the retailer needs to decide prices of $N$ types of products which are produced based on $M$ types of resources with unreplenishable initial inventory. When demand is nonparametric with some mild assumptions, \citet{miao2021network} is the first paper which proposes an algorithm with $O(\text{poly}(N,M,\ln(T))\sqrt{T})$ type of regret (in particular, $\tilde O(N^{3.5}\sqrt{T})$ plus additional high-order terms that are $o(\sqrt{T})$ with sufficiently large $T\gg N$). In this paper, we improve the previous result by proposing a primal-dual optimization algorithm which is not only more practical, but also with an improved regret of $\tilde O(N^{3.25}\sqrt{T})$ free from additional high-order terms.
 A key technical contribution of the proposed algorithm is the so-called \textit{demand balancing}, which pairs the primal solution (i.e., the price) in each time period with another price to offset the violation of complementary slackness on resource inventory constraints. Numerical experiments compared with several benchmark algorithms further illustrate the effectiveness of our algorithm. 
}
}



\date{}

\maketitle

\section{Introduction}

The price-based \emph{network revenue management (NRM)} is a classical problem in operations management, which has the following problem formulation.
Let $T$ be the time horizon, $N$ be the total number of product types and $M$ be the total number of resource types.
At the beginning of each time period $t\in[T]$, an algorithm posts a price vector $p_t\in\mathbb R_+^N$, which incurs a realized demand vector of products $y_t\in\mathbb R_+^M$,
and collects revenue $r_t=\langle p_t,y_t\rangle$ while incurring resource consumption $Ay_t \in\mathbb R_+^M$, where $A\in\mathbb{R}^{M\times N}_+$ is the resource consumption matrix.
Each resource type $j\in[M]$ has a non-replenishable initial inventory level of $C_j>0$ and a product cannot be sold if its required resources' inventory is depleted.
The objective of NRM is to maximize the expected total revenue over $T$ time periods, subject to resource inventory constraints.

In this paper we study a setting in which the expectations of realized demands are governed by a fixed but unknown \emph{demand curve} $D(p_t)$,
with demand noises being independently and identically distributed and well-behaved (see Section~\ref{sec:formulation} for detailed problem set-up and assumptions). 
Such a setting is usually referred to as \emph{blind network revenue management} because the unknown demand curve needs to be learnt on the fly,
and has been the subject of extensive studies over the past decade, pioneered by the seminal work of \cite{besbes2012blind}.
We present a high-level summary of important existing results next, which motivate our approaches and give the context for better understanding our contributions.

When the demand curve $D(\cdot)$ belongs to a certain parametric function class (e.g., linear, logistic or exponential functions with unknown parameters), 
the standard method is to use linear regression or maximum likelihood estimation to obtain an estimated demand curve $\hat D(\cdot)$ (or an upper bound of it) and then carry out optimizations.
This is for example the underlying strategy of \cite{broder2012dynamic,ferreira2018online}, some of which also have model estimation and price optimization procedures intertwined.
For non-parametric models, it is possible to estimate the \emph{entire} demand function by either using a binning approach in one dimension \citep{besbes2012blind},
or by using spline approximation in higher dimensions when the demand curve is very smooth \citep{chen2019nonparametric}.
Such approaches are not likely to be optimal for nonparametric demand curves in high dimension due to the \emph{curse of dimensionality},
as evident in the results of \citep{leimultidimensional,chen2019nonparametric,wang2021multi,chen2021nonparametric}.
More specifically, the regret must depend \emph{exponentially} on the dimension $N$ or $M$.

Another line of research for blind NRM focuses primarily on inexact optimization methods applied to certain \emph{fluid approximation} of the NRM problem,
thus circumventing the curse of dimensionality by exploiting \emph{concavity} structures in the problem satisfied by most commonly used demand models \citep{li2011pricing}.
With $T\to\infty$ and (expected) demands being stationary (meaning that $D(\cdot)$ does not change over time), it is well-known that \citep{gallego1994optimal,gallego1997multiproduct} the optimal solution
of NRM could be approximated by an optimization problem $\max_d r(d)$ subject to $Ad\leq\gamma$, with $r(d)=\langle d, D^{-1}(d)\rangle$ being the expected revenue function
and $\gamma=C/T$ being the normalized inventory vector. With $r$ being concave in $d$ and $Ad\leq \gamma$ being a polytope or halfspace feasible region, it is possible to optimize over $d$
using either first-order methods \citep{chen2019network} or cutting plane methods \citep{wang2014close,miao2021network}.
The challenge of this type of methods is obvious: because the algorithm cannot really operate in the demand domain (since the demand curve is unknown),
meaning that an algorithm cannot directly set exact demand rates but can only affect demand rates via an unknown demand curve by setting up prices,
conversions between the price and demand domains need to be carried out, which is highly non-trivial in multiple dimensions leading to sub-optimal regret \citep{chen2019network}
or high dimensional overheads \citep{miao2021network}.

Finally, the seminal work of \cite{badanidiyuru2018bandits} initialized the study of \emph{primal-dual} methods in bandits with knapsacks problems.
When learning is required, such methods are typically applied to parametric problems, because (approximate) optimal primal solutions are requested for every dual 
variable $\lambda_t$ being obtained at the beginning of each time periods.
See for example the works of \cite{agrawal2016linear,miao2021general,balseiro2020best}.
The main advantage of such methods is that the primal optimization problem could be solved \emph{without constraints}, which is feasible
even for some non-concave objectives that are well-behaved (e.g., satisfying Polyak-Lojasiewicz conditions \citep{karimi2016linear}). 
In contrast, without Lagrangian multipliers the fluid approximation in the price domain has an unknown, non-convex feasible region $AD(p)\leq\gamma$ which is much harder to solve.

For nonparametric problems, the major challenge of applying such primal-dual methods is the fact that multiple time periods are necessary
to obtain reasonably good primal solutions for a \emph{fixed} dual vector, but in the framework of \citep{badanidiyuru2018bandits} (and also all its follow-up works)
 the dual vectors must be constantly updated, interfering with the primal optimization procedure. 
 Indeed, such a challenge has deeper roots from a feasibility perspective, which we shall elaborate in much more details in the next section,
 that also motivates our solutions.

\subsection{Infeasible primal-dual methods with frequent dual updates}

To gain deeper understanding of the primal-dual framework in \citep{badanidiyuru2018bandits}, let us first summarize it at a very high level. Let $\max_a f(a)$ subject to $F(a)\leq \beta$ be
a constrained optimization problem where $a$ is the action (in our problem, $a$ denotes price vectors), and $\mL(a,\lambda)=f(a)-\langle \lambda, F(a)-\beta\rangle$ be the Lagrangian corresponding to non-negative dual vector $\lambda$, and $Q(\lambda) = \max_{a}\mL(a,\lambda)$ be the dual function. 
The work of \citep{badanidiyuru2018bandits} optimizes the constrained problem in an iterative manner: at iteration $t$, first compute $a_t=\arg\max_{a}\mL(a,\lambda_t)$ as a maximizer
of the Lagrangian at dual vector $\lambda_t$; afterwards, slack $F(a_t)-\beta$ is used to update dual vector $\lambda_{t+1}$ for the next period using an online convex optimization (OCO) method, such as online mirror descent. When $f$ or $F$
is unknown, their estimates are used to perform primal and dual updates. 

This method is very similar to \emph{dual descent}, a natural algorithm for convex optimization. Note that $Q$ is a convex function in $\lambda$ and $\nabla Q(\lambda) = \beta - F(a_\lambda^*)$
if there is a unique $a_\lambda^*$ that maximizes $\mL(\cdot,\lambda)$.
Hence, updating dual vectors $\{\lambda_t\}_t$ using online mirror descent based on $\{\nabla Q(\lambda_t)\}_t = \{\beta - F(a_t)\}_t$ naturally minimizes the dual function, which in turn
maximizes the primal function.

One key difference between BwK type algorithms and the classical dual descent algorithm, which arises because of the nature of bandit problems, is the handling of violation of primal feasible
constraints. Recall that dual descent is an \emph{infeasible} method, meaning that all primal solutions obtained on the optimization path are likely {infeasible} to the primal problem.
Furthermore, the violation of primal feasibility constraints of $a_\lambda^*$ at a sub-optimal $\lambda$ could be arbitrarily large, even if $\lambda$ is very close to $\lambda^*=\arg\min_{\lambda}Q(\lambda)$
and strong duality holds (meaning that there exists a primal feasible $a_{\lambda^*}^*$ that maximizes $\mL(\cdot,\lambda^*)$). Indeed, this is the case for LP dual problems or in general
any dual problems that are not smooth, as even the slightest amount of deviation from $\lambda^*$ leads to significantly infeasible primal solutions.

Note that, when the primal objective is concave, an averaging procedure of primal solutions could lead to approximately primal feasible solutions without additional constraints or measures.
See for example, the works of \citep{nesterov2009primal,nedic2009approximate}. Such an approach is \emph{not} applicable in our problem for two reasons.
First, the objective function is \emph{non-concave} in the price domain, which invalidates the averaging argument in the approximate feasibility analysis.
Second, approximate primal feasibility only applies to \emph{averaged} primal solutions, which cannot be extended to analyze the optimization algorithm with cumulative regret upper bounds
because the cumulative regret definition tracks all solutions at which function values are queried.

When dual descent is applied as an optimization algorithm with full information, the infeasible nature of the method is not a big deal as our purpose is only to find an optimal (thus feasible) solution in the end.
However, in bandit problems the (infeasible) primal solutions are actually implemented in sequential time periods, raising serious concerns if they are far from primal feasible.
Therefore, BwK algorithms must implement additional measures to control the violation of primal feasibility constraints, because the mere convergence of dual variables
is usually not sufficient. One of the key ideas in \citep{badanidiyuru2018bandits} is to use online convex optimization formulation that treats the slacks (gradients of the dual function) $\{\beta-F(a_t)\}_t$
as an \emph{adversarial} sequence of vectors, arguing that there exists a dual vector that limits the (cumulative) violation of primal feasibility of $\{a_t\}$. Afterwards, 
online convex optimization methods are analyzed showing that $\{\lambda_t\}$ are competitive against any fixed benchmark, even if $\{\beta-F(a_t)\}_t$ are adversarial.

Because the slacks $\{\beta_t-F(a_t)\}_t$ are treated as \emph{adversarial} in \citep{badanidiyuru2018bandits}, which is also the case in all the subsequent follow-up works
\citep{agrawal2016linear,agrawal2019bandits,miao2021general,balseiro2020best}, the ``frequent'' updates of dual vectors $\{\lambda_t\}$ are essentially unavoidable.
Otherwise, the effectiveness of the dual vectors can no longer be ensured via adversarial linear optimization arguments.
Therefore, we call this type of methods ``infeasible primal-dual methods with \emph{frequent} dual updates". The nature that dual vectors must be updated frequently has two disadvantages.
First, in some applications such as online ad allocation, dual vectors (measuring budget spending of ad-liners) cannot be updated too often due to computational constraints as the number of 
ad bidding opportunities over a given time period is huge. In applications such as dynamic pricing, it is not a good idea either to have frequent dual updates (and hence frequent primal/price updates)
as frequent changes of prices do not look good. 
Second, as is the focus of this paper, when $f$ or $F$ are nonparametric functions to be learnt, the primal optimization problem cannot be accomplished using a few number of time periods.
Because updates of dual vectors directly change the primal objectives to be maximized, analyzing such a scheme is going to be extremely difficult.
\footnote{One idea is to apply \emph{bandit convex optimization} algorithms which might allow functions to be optimized to change adversarially over time.
This has two disadvantages. First, such methods are typically very complex and most of the time impractical \citep{bubeck2021kernel}. More importantly, unlike the stationary case
where non-concave objectives could be tractable with additional constraints, in bandit convex optimization the analysis relies crucially on the first-order conditions of each function
which essentially rules out extensions to revenue functions that are not concave in price vectors.}

\subsection{Infeasible primal-dual methods with infrequent dual updates}\label{subsec:infreq_dual_update}

Motivated by the discussion in the previous section, in this paper we aim at an improved BwK algorithm by limiting the number of dual variable updates,
which facilitates primal optimization of nonparametric functions and also has potential impacts in operational practice.
To this end, we discard the perspective of treating slacks $\{\beta-F(a_t)\}_t$ as a completely adversarial sequence, 
and recover the original meanings of $\{\beta-F(a_t)\}_t$ as gradients of the dual function $Q(\cdot)$ at $\{\lambda_t\}_t$.
With suitable conditions, the dual function in our problem is strongly smooth and convex, meaning that gradient descent methods with fixed step sizes
shall have linear convergence and therefore only an $O(\log T)$ number of dual updates are required. \footnote{When $Q$ is not smooth or strongly convex, it is also possible
to limit the number of dual updates by using cutting-plane methods, such as the ellipsoid method \citep{khachiyan1979polynomial} or interior-point methods \citep{vaidya1996new}
which enjoys linear convergence for non-smooth functions. Such methods however are beyond the scope of this paper.}
This not only limits the number of dual variable updates, but also allows ample time periods to solve an unconstrained primal problem without specific parametric forms.

Because we have discarded the adversarial linear optimization perspective in \citep{badanidiyuru2018bandits},
we have to implement additional measures to control the violation of primal feasibility constraints. Because the dual function in our problem is smooth,
it is natural to conjecture that, hypothetically if $Q(\lambda)-Q(\lambda^*)=\epsilon$ is small, the violation of primal constraints at $a_\lambda^* = \arg\max_a \mL(a,\lambda)$ is also small,
because $\|F(a_\lambda^*)-F(a_{\lambda^*}^*)\|_2$ is supposedly small due to Lipschitz continuity of gradients of $Q$.
With strong duality $a_{\lambda^*}^*$ is primal feasible, and therefore $a_\lambda^*$ should be approximately feasible too.

While such an argument is intuitively correct, the regret it leads to is unfortunately sub-optimal asymptotically.
It is difficult to explain the sub-optimality in a precise mathematical manner, but here we give a sketch of the calculations.
With inexact gradients, it is possible to obtain a sub-optimal dual vector $\lambda$ with sub-optimality gap $Q(\lambda)-Q(\lambda^*)$ upper bounded by $\epsilon=O(1/\sqrt{T})$ after $T$ time periods.
With strong smoothness and convexity of $Q$, this implies that $\|F(a_\lambda^*)-F(a_{\lambda^*}^*)\|_2$ is on the order of $O(\sqrt{\epsilon})=O(T^{-1/4})$,
indicating that on average $a_\lambda^*$ violates primal feasibility constraints by an amount of $T^{-1/4}$. This shows that, the inventory of resources will be depleted
with $O(T^{3/4})$ time periods still remaining, yielding a cumulative regret of $O(T^{3/4})$ which is significantly higher than the optimal regret of $O(\sqrt{T})$.
Consequently, more refined control of primal feasibility violation in addition to near optimality of dual variables is needed, which is the main contribution of this paper
summarized in the next section.

\subsection{Summary of our contributions}\label{subsec:summary_contribution}

\textbf{Demand balancing in primal optimization.} Our first contribution, as mentioned in the previous discussion, is to solve the problem of excessive primal feasible constraint violation. The idea of our method is the following.
We shall now use $p$ to denote the price vectors and $p_\lambda^*=\arg\max_p \mL(p,\lambda)$ the price vector that maximizes the Lagrangian function at dual variable $\lambda$,
which correspond to $a$ and $a_\lambda^*$ notations in previous sections.
It can be established that if $Q(\lambda)-Q(\lambda^*)=\epsilon$ then the violation of primal feasible constraints at $p_\lambda^*$ is on the order of $O(\sqrt{\epsilon})$.
We will need to implement additional measures to bring down such violation to the order of $O(\epsilon)$ in order to achieve an optimal $O(\sqrt{T})$ overall regret.
To accomplish this, we borrow the idea of \emph{Linear Rate Control} (LRC) from the full-information re-adjustment optimization literature \citep{jasin2014reoptimization} to carefully \emph{balance} 
the demands realized at $p_\lambda^*$, so that the aggregated deviation from targeted resource consumption levels are \emph{de-biased} to a level of $O(\varepsilon)$ instead of $O(\sqrt{\varepsilon})$. Compared with the literature, a key challenge in our problem is that the demand is unknown so we cannot follow LRC directly which operates in the demand rate domain. 
Instead, we use local Taylor expansion to approximate the demand functions and adjust prices locally, which has a low-order of bias.
We shall discuss more details of our method in Section~\ref{subsec:grad-est}.


\textbf{Primal-dual gradient descent algorithm with near-optimal regret.} Based on the technique of demand balancing in primal optimization, we propose a primal-dual optimization algorithm, named \textit{Primal-Dual NRM} (PD-NRM), to solve NRM based on gradient descent. The structure of the algorithm is as follows. The time horizon $T$ is divided into epochs with exponentially growing lengths. At the beginning of each epoch $s$, there is a dual variable $\lambda_s$ fixed and the primal optimization aims to find the optimal price $p^*_{\lambda_s}$ corresponding to $\lambda_s$. By combining gradient descent and the demand balancing technique, we can show that the primal optimization can find approximately optimal prices with minimal primal feasible constraint violations, which are further used to update the dual variable $\lambda_{s+1}$ via (proximal) gradient descent. Our algorithm has the following advantages compared with the literature. First, we prove that our algorithm has a regret of $\tilde O(N^{3.25}\sqrt{T})$, which slightly improves the best bound in the literature (i.e., $\tilde O(N^{3.5}\sqrt{N})$ in \citealt{miao2021network}).
It is also free from high-order regret terms $o(\sqrt{T})$, which appears in the regret bound \citet{miao2021network} with quite heavy dimension dependency.
 Compared with \citet{miao2021network}, our algorithm is much more practical as it does not involve translating back and forth between the price and demand rate domain.
Second, our algorithm is computationally efficient even with relatively large number of products (as we use gradient descent which is efficient even with large dimension) and long time horizon (as we only need to update the policy for $O(\ln(T))$ time). Third, numerical experiments using the example in the literature (see e.g., \citealt{chen2019nonparametric}) demonstrate that our algorithm is practical and effective, as it outperforms several benchmark algorithms selected from the related literature.


\subsection{Other related works}
We first review the literature of revenue management with demand learning that we did not mention in earlier subsections. Note that this literature review is only to illustrate some most related works of our paper, and we refer the interested readers to \citet{bitran2003overview,den2015dynamic,chen2015recent} for a comprehensive summary of this field.
Let us first review the literature on revenue management with parametric demand learning. For instance, some earlier work such as \citet{araman2009dynamic,farias2010dynamic,harrison2012bayesian} study this problem using Bayesian learning and updating. When there is resource inventory constraints, \citet{den2015dynamic2} study a modified certainty-equivalent pricing policy with near-optimal regret. When demand is nonparametric, \citet{besbes2015suprising} show that using a simple linear model is sufficient to approximate the nonparametric demand which achieves near-optimal regret. If demand is further determined by some changing covariates in some nonparametric manner, \citet{chen2021nonparametric} proposes a ``divide and approximate'' learning approach with approximately optimal performance. 
Besides nonparametric and parametric demand, some paper also study the so-called semi-parametric model. For example, \citet{nambiar2019dynamic} study the learning error of a semi-linear demand model of the format $f(x_t)+\beta\cdot p$ (where $x_t$ is the demand covariates) is approximated by linear demand; similar problem is studied in \citet{bu2022context} as well.

Another related stream of literature is on solving revenue management problem with approximate heuristics when demand information is known. The seminal paper by \citet{gallego1994optimal,gallego1997multiproduct} propose a fluid approximation heuristic with a static price charged throughout the time horizon, and this heuristic is proved to have regret $O(\sqrt{T})$. Following these two papers, many follow-up research focus on improving this fluid approximation heuristic by policy re-adjustment/re-solving. Besides \citet{jasin2014reoptimization} we have mentioned earlier, another related paper is \citet{chen2016real} which propose a flexible adjustment approach with minimal price adjustments. Later on \citet{wang2020constant} show that such re-solving technique achieves constant regret, and shows that there is an optimality gap of $\Omega(\ln(T))$ between the optimal revenue and the fluid approximation revenue. Besides these paper which discuss the continuous price control, some other literature focus on the discrete decisions in revenue management such as whether the decision maker should accept or reject an order (see e.g., \citealt{cooper2002asymptotic,reiman2008asymptotically,secomandi2008analysis,jasin2012re,bumpensanti2020re}).


\section{Problem Formulation and Assumptions}\label{sec:formulation}

In the blind network revenue management problems, there are $N$ product types, $M$ resource types, and customers arrive sequentially over $T$ time periods.
At the beginning of time period $t$, a pricing algorithm posts a price vector $p_t\in[\underline p,\overline p]^N$, and when inventory is sufficient, receives realized demand $y_t\in[\underline d,\overline d]^N$. The expected demand given price $p_t$ is denoted as
\begin{equation}
\mathbb E[y_t|p_t] = D(p_t),
\label{eq:demand-model}
\end{equation}
with $D:\mathbb R_+^N\to\mathbb R_+^N$ is an \emph{unknown} demand rate function.
Afterwards, a resource vector of $A y_t\in\mathbb R_+^M$ is consumed, with $A\in\mathbb R_+^{M\times N}$ being a \emph{known} resource consumption matrix.
Each resource type $j\in[M]$ has a \emph{known} initial inventory level $C_i<\infty$. If the inventory of a resource type is completely depleted, we set $y_t=0$ almost surely (i.e., we allow a ``shut-off'' price $p_{i,t}=+\infty$ for all $i\in [N]$).
\footnote{Note that in practice, when only one resource type is depleted, products not using the depleted resource type could still be sold and the realized demand could be positive.
Nevertheless, in this paper we adopt a ``hard cutoff'' model in which demand realization is completely halted once \emph{any} resource type is depleted. This shall not affect the analysis of this paper
because with high probability the designed algorithm will only exhaust resources near the very end of the time horizon $T$.}

\subsection{Admissible policy and rewards}

We first give the rigorous definition of an \emph{admissible} policy in the sequential decision making problem.
\begin{definition}[admissible policy]
A policy $\pi$ is \emph{admissible} if it can be written as $\pi=(\pi_1,\cdots,\pi_T)$ such that $p_t\sim \pi_t(\cdot|p_1,y_1,\cdots,p_{t-1},y_{t-1})$ is measurable conditioned on the filtration of $\{p_\tau,y_\tau\}_{\tau<t}$, and furthermore satisfies $p_{t,i}=+\infty$ almost surely whenever a resource type requested by product $i$ is depleted at time $t$.
\end{definition}

To simplify our analysis, we will assume throughout this paper that $p_t y_t\in[0, B_r]$ almost surely for some constant $B_r<\infty$, which is a common assumption in bandit and contextual bandit problems
\citep{bubeck2012regret}.
Let $\Pi$ be the class of all admissible policies. For any admissible policy $\pi\in\Pi$, its expected reward is defined as
$$
\rwd(\pi) := \mathbb E\left[\sum_{t=1}^T p_ty_t\bigg| p_t\sim\pi_t(\cdot|\mathcal F_{t-1}), \;\begin{array}{ll}\mathbb E[y_t|p_t]=D(p_t)& \text{if }\sum_{\tau<t}Ay_\tau\leq C,\\ y_t=0& \text{otherwise}.\end{array}\right], 
$$
where $\mathcal F_t=\{y_\tau,p_\tau\}_{\tau<t}$.
Let $\pi^*=\arg\max_{\pi\in\Pi}\rwd(\pi)$ be the optimal policy. Then for any admissible policy $\pi\in\Pi$, its cumulative \emph{regret} is defined as
$$
\reg(\pi) := \rwd(\pi^*) - \rwd(\pi).
$$
Clearly, $\reg(\pi)\geq 0$ for all admissible policy $\pi$ by definition, and the smaller the regret the better the designed policy $\pi$ is.

\subsection{Assumptions}

Throughout this paper we impose the following assumptions on the demand function, resource consumption matrix and other problem parameters.

Our first assumption concerns the demand function $D:\mathbb R_+^N\to\mathbb R_+^N$, one of the major problem specification that is unknown in advance
and must be learnt on the fly through interactions with sequentially arriving customers.

\begin{assumption}[Assumptions on the demand curve]
The domain of $D$ is $[\underline p,\overline p]\subseteq\mathbb R_+^N$ and the image of $D$ is $D([\underline p,\overline p]^N)=\{D(p): p\in[\underline p,\overline p]\}\subseteq[\underline d,\overline d]$ for some
$0\leq \underline d<\overline d<\infty$.
Furthermore, $D$ is continuously differentiable on $[\underline p,\overline p]^N$ with bounded, Lipschitz continuous and non-degenerate Jacobians: for every $p,p'\in[\underline p,\overline p]^N$, it holds that $\|J_D(p)\|_{\op}\leq B_D$, $\sigma_{\min}(J_D(p))\geq \sigma_D>0$ and $\|J_D(p)-J_D(p')\|_\op\leq L_D\|p-p'\|_2$, where $J_D(p)=\partial D/\partial p\in\mathbb R^{N\times N}$ is the Jacobian matrix of $D$, $\sigma_{\min}(\cdot)$ indicates the smallest singular value, and $\sigma_D>0$, $B_D,L_D<\infty$ are constants.
\label{asmp:D}
\end{assumption}

Assumption \ref{asmp:D} gives quantitative specifications on the smoothness and invertibility of the multi-variate multi-value demand curve $D$, which are standard assumptions in the literature.
More specifically, Assumption \ref{asmp:D} assumes that $D$ is continuously differentiable with its Jacobian matrix $J_D$ being Lipschitz continuous in matrix operator norm, and therefore
the expected demand rates are smooth functions of the offered prices.
On the other hand, Assumption \ref{asmp:D} states that the Jacobian matrix of $D$ is non-degenerate with strictly positive least singular values,
which implies that $D$ is invertible and the inverse function $D^{-1}$ is Lipschitz continuous as well.

With Assumption \ref{asmp:D}, it is clear that $D$ is a bijection between $[\underline p,\overline p]$ and $D([\underline p,\overline p]^N)$. Let $D^{-1}$ be the inverse function of $D$. Define the expected reward function as
$$
f(p) := \langle p, D(p)\rangle \in\mathbb R_+, \;\;\;\;\;\; \forall p\in[\underline p,\overline p].
$$
Equivalently, we may also express the (expected) reward as a function of the demand rate vector $d$, with the help of the inverse function of $D$:
$$
\phi(d) := \langle d, D^{-1}(d)\rangle\in\mathbb R_+,\;\;\;\;\;\;\forall d\in D([\underline p,\overline p]^N).
$$

We impose the following conditions on the expected reward functions $f$ and $\phi$:
\begin{assumption}[Assumptions on the total revenue]
Both $f$ and $\phi$ are twice continuously differentiable, and for every $p\in[\underline p,\overline p]^N$ and $d=D(p)$, it holds that
$\|\nabla f(p)\|_2,\|\nabla^2 f(p)\|_\op\leq B_f$, $\|\nabla\phi(d)\|_2,\|\nabla^2\phi(d)\|_\op\leq B_\phi$ for some constants $B_f,B_\phi<\infty$. Furthermore, $\phi$ is strongly concave in $d$, meaning that for every $d\in D([\underline p,\overline p]^N)$,
$-\nabla^2\phi(d)\succeq \sigma_\phi I_{N\times N}$ for some constant $\sigma_\phi>0$.
\label{asmp:f-phi}
\end{assumption}

At a higher level, Assumption \ref{asmp:f-phi} assumes that the expected revenue is a smooth function of the offered prices and expected demand rates
with Lipschitz continuous gradients, which is compatible with the smoothness of the demand curve itself as characterized in Assumption \ref{asmp:D}.
Furthermore, Assumption \ref{asmp:f-phi} assumes that the expected revenue is \emph{strongly concave} in the expected demand rates.
The concavity of the expected total revenue as a function of demands rates is also a ``regular'' condition in asymptotic analysis of network revenue management problems,
as it enables approximation through size-independent concave fluid approximation problems (Section~\ref{subsec:fluid}, see also \citealt{gallego1994optimal,gallego1997multiproduct}).
The work of \cite{miao2021network} also analyzed several important demand classes and demonstrated that they satisfy Assumption \ref{asmp:f-phi} rigorously.
Note that, importantly, we do \emph{not} assume $f$ is strongly concave in the \emph{price} vector $p$, a condition frequently violated by common demand models such as the logistic demand or the exponential demand.
However, for these demand models $\phi$ is indeed strongly concave in demand rates $d$, as established in the work of \cite{li2011pricing}.

We next impose assumptions on initial inventory levels:
\begin{assumption}[Assumption on inventory levels]
There exists $\gamma_1,\cdots,\gamma_M\in[\gamma_{\min},\gamma_{\max}]$ such that $C_j=\gamma_j T$ scales linearly with time horizon $T$, where
$\gamma_{\min}>0$ and $\gamma_{\max}<\infty$ are constants.
\label{asmp:gamma}
\end{assumption}
Assumption \ref{asmp:gamma} states that the initial inventory levels of each resource type scales \emph{linearly} (albeit with different slopes) with the time horizon $T$,
which is a conventional condition when asymptotic analysis (i.e.~$T\to\infty$) is carried out in inventory systems \citep{wang2014close,besbes2012blind}.

Finally, we impose the following condition on the resource consumption matrix $A$.
\begin{assumption}[Assumption on resource consumption]
The number of resource types $M$ is smaller than or equal to the number of product types $N$. Furthermore, the resource consumption matrix $A\in\mathbb R^{M\times N}$ satisfies $\|A\|_\op\leq B_A$
and $\sigma_M(A)\geq\sigma_A$, where $\sigma_M(A)$ is the $M$th largest singular value of $A$, and $B_A<\infty$ and $\sigma_A>0$ are constants.
\label{asmp:A}
\end{assumption}

Assumption \ref{asmp:A} asserts that there are fewer (or equal number) of resource types compared to product types, and the matrix $A$ is bounded and has full row rank.
{For instance, if each resource type represents the inventory of each product itself, then we naturally have $M=N$. In other applications such as selling modular products (e.g., laptops), $M$ different modules typically can be assembled into many more types of final products, leading to the fact that $M\leq N$. Technically, we need the assumption that $A$ has full row rank (hence necessary to have $M\leq N$) to guarantee that the dual function is strongly convex, which is crucial to obtain the linear convergence of gradient descent. Unfortunately, our current analysis cannot handle the case when $M>N$. Our conjecture is that we shall separately analyze the dual variable domain, which guarantees strong convexity of the dual function, and its complement, and we leave this as a future research opportunity.
}

Apart from the assumptions in blind network revenue management listed in this section, we also impose and discuss several technical assumptions in Section~\ref{subsec:additional-asmp},
which are better stated and discussed after the introduction of the fluid approximation problem and its dual problem, and their properties.

\section{Fluid Approximation and the Dual Problem}\label{sec:fluid}

In this section, we introduce the standard size-independent fluid approximation of the network revenue management problem, and its Lagrangian dual problem. Properties of the dual problem are also discussed,
together with several additional technical assumptions facilitating our analysis.

\subsection{Fluid approximation}\label{subsec:fluid}

Consider the following fluid approximation to the network revenue management problem:
$$
\max_{d_1,\cdots,d_T\in D([\underline p,\overline p]^N)} \sum_{t=1}^T\phi(d_t)\;\;\;\;\;\;s.t.\;\; \sum_{t=1}^T Ad_t\leq C,
$$
where the $\leq$ operator is understood in an element-wise sense. Because $\phi(\cdot)$ is concave and the feasible region is linear, the fluid approximation can be reduced to
\begin{equation}
\max_{d\in D([\underline p,\overline p]^N)} \phi(d) \;\;\;\;s.t.\;\; Ad\leq \gamma,
\label{eq:fluid-d}
\end{equation}
where $\gamma = C/T$. By the bijective nature of the demand curve $D$, Eq.~(\ref{eq:fluid-d}) is also equivalent to 
\begin{equation}
\max_{p\in[\underline p,\overline p]^N} f(p)\;\;\;\;s.t.\;\; AD(p)\leq \gamma,
\label{eq:fluid-p}
\end{equation}
though neither the objective function nor the feasible region of Eq.~(\ref{eq:fluid-p}) is concave/convex.

The following proposition is standard in network revenue management, which shows that we can use the fluid approximation as a benchmark to bound the regret.
\begin{proposition}[\cite{gallego1994optimal}]
Let $d^*$ be the optimal solution to Eq.~(\ref{eq:fluid-d}). Then $T\phi(d^*)\geq \rwd(\pi^*)$, where $\pi^*=\arg\max_{\pi\in\Pi}\rwd(\pi)$ is the optimal admissible policy.
\label{prop:fluid}
\end{proposition}

\subsection{Dual problem}\label{subsec:dual}

For a price vector $p\in[\underline p,\overline p]^N$ and a dual vector $\lambda\in\mathbb R_+^M$, the Lagrangian of Eq.~(\ref{eq:fluid-p}) is 
\begin{equation}
\mL(\lambda,p) = f(p) - \langle\lambda, AD(p)-\gamma\rangle.
\label{eq:defn-Lp}
\end{equation}
Equivalently, for $d\in D([\underline p,\overline p]^N)$, define
\begin{equation}
\mH(\lambda,d)=\mL(\lambda,D^{-1}(d)) = \phi(d) - \langle\lambda, Ad-\gamma\rangle.
\label{eq:defn-Ld}
\end{equation}
Because $\mH(\lambda,d)$ is concave in $d$ and therefore has better properties, we shall deal with this Lagrangian whenever possible, though both forms of $\mL(\lambda,p)$ and $\mH(\lambda,d)$ are 
equivalent. 

For a dual vector $\lambda\in\mathbb R_+^M$, let 
\begin{equation}
p^*_\lambda := \arg\max_{p\in[\underline p,\overline p]} \mL(\lambda,p), \;\;\;\;\;\; d^*_\lambda := \arg\max_{d\in D([\underline p,\overline p]^N)}\mH(\lambda, d) = D(p^*_\lambda)
\label{defn:pd-star}
\end{equation}
be the price vector and its corresponding demand rate vector that maximizes the Lagrangian. Define the dual function
\begin{equation}
Q(\lambda) := \mL(\lambda,p^*_\lambda) = \mH(\lambda,d_\lambda^*).
\label{eq:defn-Q}
\end{equation}

Note that Eq.~(\ref{eq:fluid-d}) satisfies the Slater's condition because the objective is concave, and all constraints are linear. Subsequently, the duality gap of this problem is zero:
\begin{lemma}[strong duality]
Let $d^*$ be the optimal solution to Eq.~(\ref{eq:fluid-d}). Then there exists $\lambda^*\in\mathbb R_+^M$ such that
$Q(\lambda^*)=\mH(\lambda^*,d^*)=\phi(d^*)$, $\nabla\phi(d^*)=A^\top\lambda^*$, $\langle\lambda^*, Ad^*-\gamma\rangle = 0$ and furthermore $Q(\lambda^*)\leq Q(\lambda)$ for any other $\lambda\in\mathbb R_+^M$.
\label{lem:strong-duality}
\end{lemma}

Our second lemma shows that $Q$ is twice continuously differentiable and strongly convex, whose proof can be found in Section \ref{sec:proof_sec3} in the appendix.
\begin{lemma}
Suppose for some $\lambda\in\mathbb R_+^N$, $d_\lambda^*$ lies in the interior of $D([\underline d,\overline d]^N)$.
Then $Q$ is twice continuously differentiable at $\lambda$ with
\begin{equation*}
\nabla Q(\lambda) = \gamma - A d_\lambda^*,\;\;\;\;
\nabla^2 Q(\lambda) = -A\big[\nabla^2 \phi(d_\lambda^*)\big]^{-1}A^\top,\;\;\;\;
\partial d_\lambda^*/\partial\lambda = \big[\nabla^2 \phi(d_\lambda^*)\big]^{-1}A^\top.
\end{equation*}
Furthermore, by Assumptions \ref{asmp:f-phi} and \ref{asmp:A}, it holds that $\nabla^2 Q(\lambda)\succeq \sigma_A^2/B_\phi I$ and $\|\nabla^2 Q(\lambda)\|_\op\leq B_A^2/\sigma_\phi I$, indicating that $Q$ is strongly convex.
\label{lem:strong-convexity}
\end{lemma}

\subsection{Additional technical assumptions}\label{subsec:additional-asmp}

We impose the following additional technical assumption on the primal and dual space.
\begin{assumption}[Assumption on primal dual space]
There are known convex compact sets $\Lambda\subseteq\mathbb R_+^M$ and $P\subseteq[\underline p,\overline p]^N$ and constants $\bar\lambda,\bar\rho<\infty$, $\underline\rho>0$, such that the following hold:
\begin{enumerate}
\item $\lambda^*\in\Lambda$, $p^*\in P$;
\item For every $\lambda\in\Lambda$, $p_\lambda^*\in P$ and $d_\lambda^*$ lies in the interior of $D([\underline p,\overline p]^N)$;
\item For every $p\in P$, $\|p-p^*\|_2\leq\bar\rho$; for every $\lambda\in\Lambda$, $\|\lambda\|_2\leq\bar\lambda$;
\item For every $p$ with $\|p-p^*\|_2\leq \chi\bar\rho$, $p_i\pm \underline \rho\in[\underline p,\overline p]$ for all $i\in[N]$, where $\chi=\sqrt{\frac{2B_\phi}{\sigma_\phi}}\frac{B_D}{\sigma_D}$.
\end{enumerate}
\label{asmp:Lambda}
\end{assumption}

Although we call all the conditions above as assumptions, we shall note that they are indeed about problem parameter designs. For instance, for the first two points in the assumption, we can always make the price range $[\underline p,\overline p]$ large enough such that $p^*_\lambda$  is in the interior of the domain (hence so does $d^*_\lambda$) for any $\lambda\in \Lambda$. On the other hand, $\Lambda$ and $P$ are just some reasonable guess of the range of $\lambda^*$ and $p^*$ respectively (as if $\lambda_j$ is too large, any products using resource $j$ will have their prices equal to $+\infty$, implying $\lambda^*$ have to be bounded). 

\section{Algorithm Design and Theoretical Results}\label{sec:algorithm}

In this section, we introduce the details of our proposed Algorithm PD-NRM, which consists of Algorithms \ref{alg:grad-est}-\ref{alg:dual-opt}. Before we go into details in the next several subsections, let us first introduce the high-level ideas of the proposed algorithm. 

\begin{figure}[t]
	\centering
	\label{fig:main_alg}
	\includegraphics[scale=0.55]{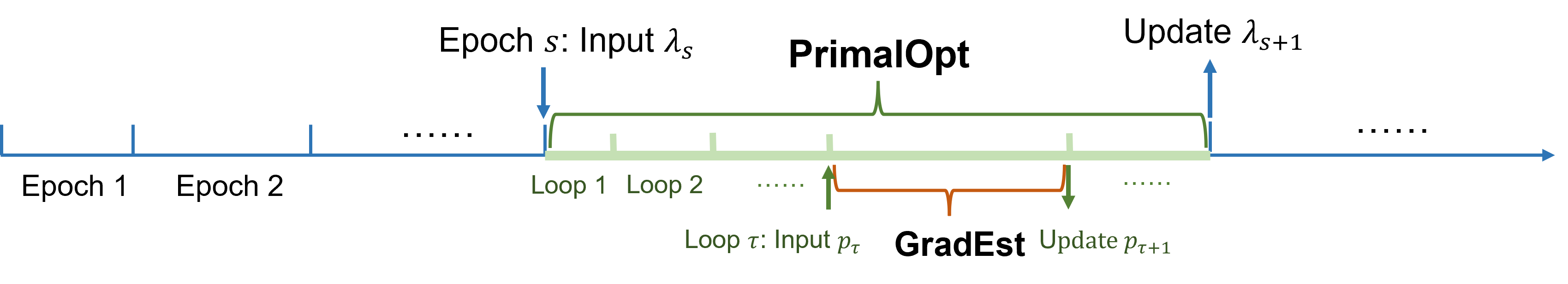}
		\caption{Structure of Algorithm PD-NRM.}
\end{figure}

As we discussed in the introduction, our algorithm is based on a primal-dual framework with infrequent update for both dual and primal optimization. 
More specifically, the time horizon $T$ is divided into epochs with exponentially increasing lengths, each of which is used to update the dual variable. Besides the dual optimization which is our main algorithm, there are two sub-routines: \textsc{PrimalOpt} (which aims to calculate $p^*_\lambda$ and gradient of dual function $\nabla Q(\lambda)$ for any $\lambda$) and \textsc{GradEst} (which is used to estimate the gradient of Lagrangian $\mL(\lambda,p)$ for any $\lambda,p$ and do the demand balancing). We refer to Figure \ref{fig:main_alg} for a graphic representation of our Algorithm PD-NRM. 

From Figure \ref{fig:main_alg}, we can see the structure of Algorithm PD-NRM as well as how sub-routines are applied. In particular, for each epoch $s$, \textsc{PrimalOpt} takes the dual variable $\lambda_s$ as the input and its output helps to update the new dual variable $\lambda_{s+1}$, which will be used for the next epoch. To implement \textsc{PrimalOpt}, epoch $s$ is further divided into smaller epochs (to avoid confusion with the outer epoch, we call them loops) with exponentially increasing lengths again. In each loop $\tau$, $p_\tau$ is the current price (for estimating $p^*_{\lambda_s}$) which is an input of the sub-routine \textsc{GradEst}. \textsc{GradEst} runs through loop $\tau$ to estimate the gradient of the Lagrangian $\mL(\lambda_s,p_s)$ and balance the demand, and thus $p_{\tau+1}$ is updated at the end of loop $\tau$. 

In the next several subsections, we will discuss the details of each algorithm and their theoretical results. Since there are many learning parameters in the algorithms and technical lemmas, we summarize them in Table \ref{tab:constants} including where they are specified, which parameters they depend on, and their asymptotic orders with respect to problem size $N$ and $T$.

\begin{table}[h!]
\centering
\caption{Summary of constants. (Polynomial) dependency on problem parameters other than $N,M$ and $T$, such as $\underline p,\overline p,\underline d,\overline d,B_D,\sigma_D,L_D,B_f,B_\phi,\sigma_\phi,B_A,\sigma_A,\gamma_{\max},\bar\lambda,\bar\rho,\underline\rho$, are omitted.}
\label{tab:constants}
\begin{tabular}{c|ccc}
\hline
Constants& Specified in ...& Depending on ...& Asymptotic order\\
\hline
$n_0$& Lemma \ref{lem:primal-opt}& $\kappa_4$& $O(N^4\ln^2(NT))$ \\
$\kappa_1$& Algorithm \ref{alg:primal-opt}& $n_0$& $O(N\sqrt{\ln(NT)})$ \\
$\kappa_2$& Algorithm \ref{alg:primal-opt}& $\kappa_5$& $O(N^{2.75}\ln^{1.5}(NT) + N^2\ln^3(NT))$\\
$\kappa_3$& Algorithm \ref{alg:grad-est}& $\kappa_1$& $O(N^{2.5}\sqrt{\ln(NT)})$\\ 
$\kappa_4$& Lemma \ref{lem:primal-opt}& N/A& $O(N\sqrt{\ln(NT)})$\\
$\kappa_5$& Lemma \ref{lem:dual-opt}&$\kappa_1,\kappa_3,\kappa_6,\mu$&$O(N^{5.5}\ln^3 (NT)+N^4\ln^6(NT))$\\
$\kappa_6$& Lemma \ref{lem:dual-opt}& $\mu$& $O(\sqrt{N})$ \\
$\eta_1$& Lemmas \ref{lem:primal-opt}, \ref{lem:dual-opt}& N/A& $O(1)$ \\
$\eta_2$& Lemma \ref{lem:dual-opt}& N/A& $O(1)$\\
$\mu$& Lemma \ref{lem:dual-opt}& N/A& $O(1)$\\
\hline
\end{tabular}
\end{table}

\begin{remark}
{
Although Table \ref{tab:constants} and the corresponding lemmas outline how to choose the learning parameters from a theoretical perspective, 
directly implementing such parameter values might be challenging or even impractical due to the numerical constants omitted in asymptotic notations.
Fortunately, we show later on in our numerical experiments in Section \ref{sec:num_exp} that some simple, ad-hoc parameter tuning is already sufficient to achieve reasonable performances,
especially compared with other baseline and competing algorithms.
This also shows that practical effectiveness and robustness of our proposed algorithm based on first-order methods.
On the other hand, it is in general a challenging task to tune algorithm parameters in a data-driven way for machine learning and bandit learning algorithms for optimal practical performances,
which we refer the readers to the works of \citealt{bardenet2013collaborative,frazier2018tutorial,yang2020hyperparameter} and would be an excellent topic for future research.
}
\end{remark}

\subsection{Gradient estimation and demand balancing}\label{subsec:grad-est}

Our first sub-routine is presented in Algorithm \ref{alg:grad-est}, which accomplishes two goals simultaneously with an input of price vector $p$, dual vector $\lambda$, sample budget $n$ (i.e., the length of the sub-routine),
returns an estimated gradient $\partial_p\mL(\lambda,p)$ and also performs \emph{demand balancing} at $p$.

{The main idea of the algorithm is as follows. First, the length of the sub-routine $n$ is divided into two phases with equal length $n/2$. In the first phase, we essentially uniformly sample around $p$ in order to estimate $D(p),J_D(p),\nabla f(p)$ simultaneously. Such estimation method was first studied in the seminal work \citet{nemirovskij1983problem,flaxman2004online}, and has been applied in several works including \citet{hazan2014bandit,besbes2015non,chen2019network,chen2019nonstationary,miao2021network}. However, simply implementing $p$ might violate the primal feasible constraints by too much, as discussed in Section \ref{subsec:infreq_dual_update}, and this motivates our demand balancing which leads to a modified price $\tilde p$. The balanced price $\tilde p$ is then applied in the second phase for another $n/2$ time periods.
}

\begin{algorithm}[t]
\caption{Gradient estimation and demand balancing}
\label{alg:grad-est}
\begin{algorithmic}[1]
\Function{GradEst}{$p,\lambda,n,\kappa_1,\kappa_2$}
	\State$u\gets\sqrt{N}/\sqrt[4]{n}\wedge \inf\{\delta>0: p_i\pm \delta\in[\underline p,\overline p],\forall i\}$, $\kappa_3\gets 4\bar dL_D\kappa_1\sqrt{N^3\ln(2NT)} + 3L_D\kappa_1^2$;\label{line:kappa-3}
	\For{$i=1,2,\cdots,N$}
		\State Commit to $p_{i,+} = p + ue_i$ for $n/(4N)$ time periods and let $d_{i,+}$ be the average demand;
		\State Commit to $p_{i,-}=p-ue_i$ for $n/(4N)$ time periods and let $d_{i,-}$ be the average demand;
	\EndFor
	\State $\hat D(p)\gets \frac{1}{2N}\sum_{i=1}^{N} (d_{i,+}+d_{i,-})$;
	\State $\hat J_D(p)\gets (\hat\partial_1 D(p),\cdots,\hat\partial_N D(p))$, where $\hat\partial_i D(p)=(d_{i,+}-d_{i,-})/(2u)$;
	\State $\hat \nabla f(p)\gets (\hat\partial_1 f(p),\cdots,\hat\partial_N f(p))$, where $\hat\partial_i f(p) = (\langle p_{i,+},d_{i,+}\rangle - \langle p_{i,-},d_{i,-}\rangle)/(2u)$;
	\State\label{line:local_taylor} Find $\tilde p\in[\underline p,\overline p]^N$ that satisfies the following (if no such $\tilde p$ exists, set $\tilde p\gets p$):\label{line:tilde-p}
	\begin{enumerate}
	\item\label{line:local_taylor1} For every $i\in[N]$, $|\tilde p_i-p_i|\leq \kappa_1n^{-1/4}$;
	\item For every $j\in[M]$, $\langle a_j, \hat D(p) + \frac{1}{2}\hat J_D(p)(\tilde p-p)\rangle \leq \gamma_j + \kappa_3/\sqrt{n}$;
	\item\label{line:local_taylor3} For every $j\in[M]$, $\langle a_j, \hat D(p) + \frac{1}{2}\hat J_D(p)(\tilde p-p)\rangle \geq \gamma_j - \kappa_2/((1\wedge\lambda_j)\sqrt{n})-\kappa_3/\sqrt{n}$;
	\end{enumerate}
	\State Commit to $\tilde p$ for the next $n/2$ time periods; return $\hat D(p)$ and $\hat J_D(p)$;
\EndFunction
\end{algorithmic}
\end{algorithm}

{Now let us discuss the details of demand balancing and why we need it. By the algorithm design of Algorithm PD-NRM, whenever we are implementing \textsc{GradEst}, we have $\|p-p^*\|_\infty=\tilde O(n^{-1/4})$ (with high probability), implying that $\mL(\lambda,p^*)-\mL(\lambda,p)=\tilde O(n^{-1/2})$ by strong smoothness of the revenue and the fact that $p^*$ is an interior optimizer. Such difference in the revenue function will eventually lead to an eventual regret of $\tilde O(\sqrt{T})$ (ignoring factor of $N$) which is exactly what we want, if we disregard the possible violation complementary slackness of inventory constraints. Unfortunately, when we take into account such violation it is challenging to obtain $\tilde O(\sqrt{T})$ regret, as explained below:
\begin{enumerate}
\item When resource is scarce (i.e., $AD(p^*)=\gamma$), we need $\|AD(p)-\gamma\|_\infty=\|AD(p)-AD(p^*)\|_\infty\leq \|A\|_{op}\|D(p)-D(p^*)\|_\infty=\tilde O(n^{-1/2})$ (if we only have the first phase in \textsc{GradEst} as we want $\|n\cdot (D(p)-D(p^*))\|_\infty=\tilde O(\sqrt{n})$) in order to have the cumulative violation of the primal feasible constraints at most $\tilde O(\sqrt{T})$,
so that the algorithm will not run out of inventory until $\tilde O(\sqrt{T})$ time periods remain. However, this target is not possible because $\|D(p)-D(p^*)\|_\infty=\tilde O(n^{-1/4})$ by Lipschitz continuity of $D(\cdot)$ and the fact that $\|p-p^*\|_\infty=\tilde O(n^{-1/4})$.
This means that, the algorithm could potentially run out of inventory with approximately $T^{3/4}$ time periods still remaining, leading to a regret of $\tilde O(T^{3/4})$.
\item When the dual variable $\lambda_j$ is non-zero, the total rewards collected by the algorithm might suffer if the actual resource consumption is far \emph{below} the average inventory level $\gamma_j$ as well.
In such cases, though the algorithm is unlikely to run out of inventory (at least not on resource type $j$), the $\lambda_j([AD(p)]_j-\gamma_j)$ term in the Lagrangian function would be large
meaning that $f(p)$ is actually much smaller than $\mL(\lambda,p)$. In other words, by consuming too little inventory for resource type $j$ whose inventory constraint in the fluid approximation
is supposed to be binding, the regret of the algorithm also increases.
\end{enumerate}

\begin{figure}[t]
\centering
\includegraphics[width=0.9\textwidth]{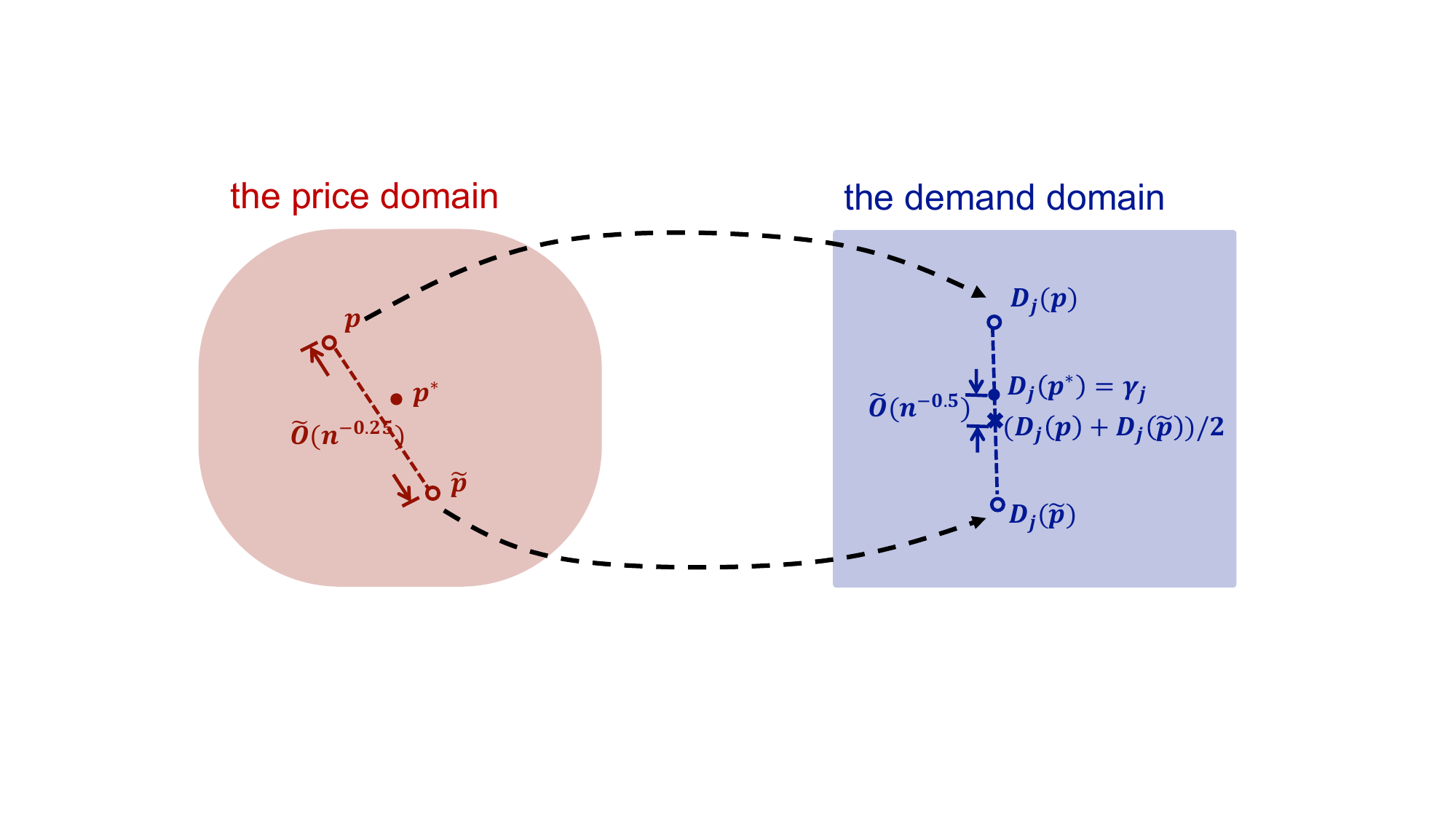}
\label{fig:demand-balancing}
\caption{Graphical illustration of the demand balancing procedure.}
{\small This figure depicts the demand balancing step on Line \ref{line:tilde-p} of Algorithm \ref{alg:grad-est} in both the price domain and the demand domain (for resource type $j$ whose inventory constraint
in the fluid-approximation is binding).
Note that in the price domain, the distance between $p$ and $\tilde p$ is on the order of $\tilde O(n^{-1/4})$ which is the order of the estimation error between $p$ and $p^*$.
With the balancing step, the \emph{average} consumption of resource type $j$ between $p$ and $\tilde p$ will be on the lower order of $\tilde O(n^{-1/2})$,
thanks to the local Taylor expansion step that eliminates first-order error terms between $p,\tilde p$ and $p^*$.}
\end{figure}

 To resolve the above-mentioned challenges, we propose the following idea which we call \emph{demand balancing}, with a graphical illustration given in Figure \ref{fig:demand-balancing}.
 The key idea is to ``de-bias'' the price offering $p$ so that the consumption of type-$j$ resource approximately matches its normalized inventory level $\gamma_j$, provided that $\lambda_j$ is
 not too small.
Instead of sampling around $p$ for the whole time horizon of $n$ periods, in the second phase, we can try to find another price $\tilde p$ such that for any $j\in [M]$,
\begin{equation}\label{eq:demand_balance_final}
	\frac{\langle a_j,D(p)+D(\tilde p)\rangle}{2}\in [\gamma_j-\tilde O(n^{-1/2}+n^{-1/2}/\lambda_j),\gamma_j+\tilde O(n^{-1/2})]	
\end{equation}
and apply $\tilde p$ for another $n/2$ time periods. Then we have the total primal violation in two phases is roughly equal to $\|(n/2\cdot(AD(p)-\gamma)+n/2\cdot(AD(\tilde p)-\gamma))^+\|_\infty=\tilde O(\sqrt{n})$ which is exactly what we want to achieve. However, since we only have access to the estimated $\hat D(p)$ and  do not know $D(\tilde p)$ for any other price $\tilde p$ either, we need to apply certain (estimated) local Taylor expansion of the demand curve in order to achieve (\ref{eq:demand_balance_final}), which is exactly the procedure in Line \ref{line:local_taylor} of Algorithm \ref{alg:grad-est}.
	
In spirit, our idea of demand balancing is similar to the \textit{Linear Rate Control} (LRC) in \citet{jasin2014reoptimization}, which is to re-adjust the fluid approximation policy under the full demand information. According to \citet{jasin2014reoptimization}, 
LRC re-adjusts the fluid price $p^*$ in order to de-bias the ``error'' of previous time periods, which means that whenever the realized cumulative demand of previous time periods goes up/down compared with the expected fluid demand $d^*$, the price is adjusted up/down correspondingly. In our case, demand balancing aims to de-bias the ``error'' of $p$ in order to guarantee that on average the primal feasible constraints are approximately satisfied. As a result, besides unknown demand information, a key difference in demand balancing is that the error to de-bias does not come from the deviation between realized and expected demand, but from the estimation error of $p^*$ (hence $d^*$).

}


Next we introduce two technical lemmas which show the effectiveness of \textsc{GradEst}. First, the following lemma upper bounds the estimation error of the expected demand rates $D$ and its Jacobian $J_D$.
\begin{lemma}
With probability $1-\tilde O(T^{-1})$ the following hold:
\begin{enumerate}
\item For every $i\in[N]$, $|\hat D_i(p)-D_i(p)|\leq \bar d\sqrt{{2N\ln(2NT)}} n^{-1/2} + 2L_Du^2$; furthermore, if $u=\sqrt{N}/\sqrt[4]{n}$ then $|\hat D_i(p)-D_i(p)|\leq 2\bar d\sqrt{N}(\sqrt{\ln(2 NT)}+L_D\sqrt{N})n^{-1/2}$;
\item For every $i\in[N]$, $\|\hat\partial_i D(p)-\partial_i D(p)\|_2 \leq 0.5L_D u + 2\bar dN u^{-1}\sqrt{\ln(2NT)}n^{-1/2}$; furthermore, if $u=\sqrt{N}/\sqrt[4]{n}$ then $\|\hat\partial_i D(p)-\partial_i D(p)\|_2\leq (2\bar d\sqrt{\ln(2NT)}+L_D)\sqrt{N} n^{-1/4}$;
\item $\|\hat\nabla f(p)-\nabla f(p)\|_2\leq (B_f u+B_r\sqrt{N}u^{-1})\sqrt{2N\ln(2NT)}$; furthermore, if $u=\sqrt{N}/\sqrt[4]{n}$ then $\|\hat\nabla f(p)-\nabla f(p)\|_2\leq (B_f\sqrt{N}+B_r)\sqrt{2N\ln(2NT)} n^{-1/4}$.
\end{enumerate}
\label{lem:est-D-JD}
\end{lemma}

Our next lemma shows that, with high probability any feasible solution $\tilde p$ to the problem on Line \ref{line:tilde-p} ``balances'' the demand at price $p$ towards the normalized target inventory level $\gamma$. It also shows that, when $D(p)$ itself is not too far away from $\gamma_j$, feasibility of Line \ref{line:tilde-p} is guaranteed.
\begin{lemma}
Suppose the \textsc{GradEst} sub-routine is executed with $\kappa_1,\kappa_2\geq 1$, and that $p_i\pm \sqrt{N}/\sqrt[4]{n}\in[\underline p,\overline p]$ for all $i\in[N]$. Jointly with the success events in Lemma \ref{lem:est-D-JD}, which occurs with probability $1-\tilde O(T^{-1})$, the following hold:
\begin{enumerate}
\item For any $\tilde p$ feasible to Line \ref{line:tilde-p}, it holds hat
$$
\gamma_j - \frac{\kappa_2}{(1\wedge\lambda_j)\sqrt{n}} -\frac{2\kappa_3}{\sqrt{n}}  \leq \frac{\langle a_j, D(p)+D(\tilde p)\rangle}{2} \leq \gamma_j + \frac{2\kappa_3}{\sqrt{n}};
$$
\item If $\|p-p^*\|_{\infty}\leq \kappa_1/(2\sqrt[4]{n})$ and $Q(\lambda)-Q(\lambda^*)\leq \kappa_2/\sqrt{n}$, where $p^*$ is the optimal fluid solution and $\lambda^*=\arg\max_{\lambda\in\mathbb R_+^M}Q(\lambda)$ is the optimal dual vector, then the feasible region described in Line \ref{line:tilde-p} is non-empty.
\end{enumerate}
\label{lem:feasibility}
\end{lemma}

{
The second point of Lemma \ref{lem:feasibility} basically says that whenever $p$ and $\lambda$ are close to $p^*$ and $\lambda^*$, the feasibility problem in Line \ref{line:tilde-p} is always feasible, which leads to the property in the first point of Lemma \ref{lem:feasibility}. As we discussed earlier, the conditions regarding $p$ and $\lambda$ in the second property can be proved to hold with high probability whenever we invoke sub-routine \textsc{GradEst}. Therefore, we can almost guarantee to find $\tilde p$ satisfying the two inequalities in the first point. Regarding these two inequalities, we can immediately see that 
\[
\frac{\langle a_j, D(p)+D(\tilde p)\rangle}{2} \leq \gamma_j + \frac{2\kappa_3}{\sqrt{n}}
\]
approximately guarantees the primal feasible constraint of resource $j$. On the other hand, the other inequality
\[
\gamma_j - \frac{\kappa_2}{(1\wedge\lambda_j)\sqrt{n}} -\frac{2\kappa_3}{\sqrt{n}}  \leq \frac{\langle a_j, D(p)+D(\tilde p)\rangle}{2}
\]
is to guarantee the approximate strong duality of the algorithm.
Notice that, in addition to the $2\kappa_3/\sqrt{n}$ term that takes care of the stochastic error, there is an additional term that scales inverse linearly with $\lambda_j$
on the lower bound side of the constraint.
Intuitively, this enforces approximate complementary slackness and allows the resource consumption to be significantly lower than the initial inventory level $\gamma_j$
provided that $\lambda_j$ is close to zero, indicating a resource constraint that may not be binding at optimality. (At exact optimality, $\lambda_j^*$ must be exactly zero if the constraint
on resource type $j$ is not binding, but the lower bound in Lemma \ref{lem:feasibility} allows non-zero but small dual variable values to accommodate for errors in dual optimization.)

}


\subsection{Primal optimization}\label{subsec:primal-opt}

When the dual variable $\lambda\in\Lambda\subseteq\mathbb R_+^M$ is given, the primal optimization procedure optimizes the penalized Lagrangian function 
$$
\mL(\lambda,p) = f(p) - \langle\lambda, AD(p)-\gamma\rangle
$$
over $p\in[\underline p,\overline p]^N$. Note that $\mL(\lambda,\cdot)$ is not necessarily concave in $p$; however, it satisfies the Polyak-Lojasiewicz (PL) inequality \citep{karimi2016linear} and is equipped with a quadratic decay property, as established in the following lemma:
\begin{lemma}
For every $\lambda\in\Lambda$ let $p_\lambda^*=\arg\max_p\mL(\lambda,p)$, which belongs to $[0.75\underline p+0.25\overline p,0.25\underline p+0.75\overline p]^N$ thanks to Assumption \ref{asmp:Lambda}. Then for any $p\in[\underline p,\overline p]^N$, it holds that
$$
\frac{1}{2}\|\nabla_p\mL(\lambda,p)\|_2^2 \geq \sigma_D^2\sigma_\phi\big[\mL(\lambda,p_\lambda^*)-\mL(\lambda,p)\big].
$$
Furthermore, for any $p$ it holds that
$$
\mL(\lambda,p) \leq \mL(\lambda, p_\lambda^*) - \frac{\sigma_\phi\sigma_D^2}{2}\|p-p_\lambda^*\|_2^2.
$$
\label{lem:pl}
\end{lemma}

\begin{algorithm}[t]
\caption{The primal optimization procedure given dual variable $\lambda$}
\label{alg:primal-opt}
\begin{algorithmic}[1]
\Function{PrimalOpt}{$\lambda,\bar\varepsilon,\eta, n_0,\kappa_5$}
	\State Initialize: $p_0\in P$;
	\For{$\tau=0,1,2,\cdots$ until $n_\tau> \kappa_5/\bar\varepsilon^2$}
		\State $n_\tau\gets \lceil (1-\eta\sigma_D^2\sigma_\phi/2)^{-2\tau} n_0\rceil$, $\kappa_1=\sqrt{\frac{8B_\phi B_D^2\bar\rho^2}{\sigma_\phi\sigma_D^2}}\sqrt[4]{n_0}$ and $\kappa_2=\sqrt{\kappa_5}$;\label{line:kappa-1}
		\State $\hat D(p_\tau),\hat J_D(p_\tau),\hat\nabla f(p)\gets\textsc{GradEst}(p_\tau,\lambda,n_\tau,\kappa_1,\kappa_2)$;
		\State $\hat\nabla_p\mL(\lambda,p_\tau) \gets \hat\nabla f(p) - \hat J_D(p_\tau)^\top A^\top\lambda$;
		\State $p_{\tau+1}\gets p_\tau + \eta\hat\nabla_p\mL(\lambda,p_\tau)$; 
	\EndFor
	\State \textbf{return} $p_\tau$, $\hat D(p_\tau)$;
\EndFunction
\end{algorithmic}
\end{algorithm}

Because of Lemma \ref{lem:pl}, we can essentially use gradient descent to optimize $\mL(\cdot,\lambda)$, and the proposed primal optimization method is presented in Algorithm \ref{alg:primal-opt}. At a higher level, the algorithm carries out inexact gradient descent on $\mL(\lambda,\cdot)$, with gradients 
estimated by the \textsc{GradEst} sub-routine in Algorithm \ref{alg:grad-est} with carefully chosen sample budget parameters $n$ as well as demand balancing being performed in the meantime.
{The effectiveness of Algorithm \ref{alg:primal-opt} is summarized in the next lemma, which upper bounds the (cumulative) errors of both Lagrangian objectives and violation of complementary slackness
at all resource constraints.}

\begin{lemma}
Suppose \textsc{PrimalOpt} is executed with $\lambda\in\Lambda$, $\bar\varepsilon\geq Q(\lambda)-Q(\lambda^*)$, $\eta=1/8(B_f+B_AB_J\bar\lambda)$, $n_0=\frac{(1+B_A\bar\lambda)^4\kappa_4^4\ln^2 T}{B_\phi^2B_D^4\bar\rho^4}\vee \frac{N^2}{\underline\rho^4}$ with $\kappa_4 = 2\bar d(L_D\sqrt{N}\vee(B_f\sqrt{N}+B_r))\sqrt{N\ln(2NT)}$, and $\kappa_5\geq 1$. Then with probability $1-\tilde O(T^{-1})$ it holds for every $\tau$ that
\begin{align*}
\mL(\lambda,p_\lambda^*)-\mL(\lambda,p_\tau) &\leq  B_\phi B_D^2\bar\rho^2\sqrt{\frac{n_0}{n_\tau}};\\
\|p_\tau-p_\lambda^*\|_2 &\leq \sqrt{\frac{2B_\phi B_D^2\bar\rho^2}{\sigma_\phi\sigma_D^2}}\sqrt[4]{\frac{n_0}{n_\tau}}\leq \frac{\kappa_1}{2\sqrt[4]{n_\tau}};\\
p_{\tau,i}\pm\underline\rho &\in [\underline p,\overline p], \;\;\;\;\forall i\in[n];\\
\frac{f(p_\tau)+f(\tilde p_\tau)}{2} &\geq f(p^*) - \frac{1.2\kappa_1^2 (B_f+B_AB_J\bar\lambda) N + (2\kappa_3+\sqrt{\kappa_5})\bar\lambda\sqrt{N}}{\sqrt{n_\tau}};\\
\frac{\langle a_j, D(p_\tau)+D(\tilde p_\tau)\rangle}{2} &\leq \gamma_j + \frac{\kappa_3}{\sqrt{n_\tau}}, \;\;\;\;\;\forall j\in[M],
\end{align*}
where $\underline\rho>0$ is the constant specified in Assumption \ref{asmp:Lambda}, $\kappa_1$ is the constant specified on Line \ref{line:kappa-1} in Algorithm \ref{alg:primal-opt}; $\kappa_3$ is the constant specified on Line \ref{line:kappa-3} in Algorithm \ref{alg:grad-est}, and $\tilde p_\tau$ is the demand balancing price vector produced on Line \ref{line:tilde-p} in Algorithm \ref{alg:grad-est}.
\label{lem:primal-opt}

\end{lemma}

{
There are several implications of Lemma \ref{lem:primal-opt}. When the dual variable $\lambda$ is sufficiently close to $\lambda^*$ (as reflected by $\bar\varepsilon\geq Q(\lambda)-Q(\lambda^*)$, which can be proved to hold with high probability), the first three conclusions in Lemma \ref{lem:primal-opt} show that $p_\tau$ is close to $p_\lambda$, depending on how long the loop is, for every loop $\tau$. As we have seen in Lemma \ref{lem:feasibility}, this proximity of price, together with $\bar\varepsilon\geq Q(\lambda)-Q(\lambda^*)$, guarantees that there exists some feasible $\tilde p$ satisfying the two inequalities in Lemma \ref{lem:feasibility}. Then the last two conclusions in Lemma \ref{lem:primal-opt} are mainly from Lemma \ref{lem:feasibility}, which guarantee that in each loop $\tau$ the revenue of Algorithm \ref{alg:primal-opt} is approximately optimal (with gap at most $\tilde O(n_{\tau}^{-1/2})$) and the primal feasible constraints are approximately satisfied (with violation at most $\tilde O(n_{\tau}^{-1/2})$) as well.
}

\subsection{Dual optimization and the main algorithm}

{In the previous two sub-sections, we introduced the two sub-routines \textsc{GradEst} and \textsc{PrimalOpt} which are used to solve the primal optimization to find $p^*_{\lambda_s}$ in epoch $s$ (with dual variable $\lambda_s$). In this subsection, we present a dual optimization \textsc{DualOpt} to minimize $Q(\lambda)$ described in Algorithm \ref{alg:dual-opt}. Together with the previous two sub-routines, they form Algorithm PD-NRM. For \textsc{DualOpt}, on the dual space $\Lambda\subseteq\mathbb R_+^M$ we use the accelerated proximal gradient method with inexact gradient oracles to carry out updates of dual vectors $\lambda$.
}

\begin{algorithm}[t]
\caption{Dual optimization via inexact projected gradient descent}
\label{alg:dual-opt}
\begin{algorithmic}[1]
\Function{DualOpt}{$T,\eta_1,\eta_2,\mu,\kappa_5,\kappa_6,n_0$}
	\State Initialize: arbitrary $\lambda_0\in\Lambda$, $s\gets 0$;
	\While{$T$ time periods are not elapsed}
		\State $\hat p(\lambda_s), \hat D(\hat p(\lambda_s))\gets\textsc{PrimalOpt}(\lambda_{s}, (1+\mu\eta_2)^{-s/2}\kappa_6, \eta_1, n_0,\kappa_5)$; \Comment{Primal optimization}
		\State $\hat\nabla Q(\lambda_s)\gets \gamma - A\hat D(\hat p(\lambda_s))$;\Comment{Inexact gradient estimation}
		\State $\hat\nabla h(\lambda_s) \gets \hat\nabla Q(\lambda_s) - \mu\lambda_s$;
		\State $\lambda_{s+1} \gets \arg\min_{\lambda\in\Lambda}\{\langle\hat\nabla h(\lambda_s),\lambda\rangle + \frac{\mu}{2}\|\lambda\|_2^2 + \frac{1}{2\eta_2}\|\lambda-\lambda_s\|_2^2\}$; \Comment{Projected gradient descent}\label{line:pgd-dual}
	\EndWhile 
\EndFunction
\end{algorithmic}
\end{algorithm}

Algorithm \ref{alg:dual-opt} describes a projected gradient descent method with inexact gradient estimates to optimize the dual problem, which serves as the entry point of the entire
algorithm and is the outermost loop of the whole algorithm.
More specifically, with the dual function $Q$ being both strongly convex and smooth thanks to Lemma \ref{lem:strong-convexity}, and the feasible region of the dual vectors is $\Lambda\subseteq\mathbb R_+^M$,
Algorithm \ref{alg:dual-opt} carries out a projected gradient descent with constant step sizes $\eta_2$.
For each iteration in the projected gradient descent, the gradient $\nabla Q(\lambda_t)$ is the slack at the primal solution that optimizes the Lagrangian function with dual variables being $\lambda_t$,
which is obtained by invoking the \textsc{PrimalOpt} procedure described in previous sections.

{Next lemma is the last piece we need to prove the regret of Algorithm PD-NRM. In particular, it shows that for every epoch $s$, $\lambda_s$ is a sufficiently good estimator of $\lambda^*$ in that $Q(\lambda_s)-Q(\lambda^*)\leq \bar\varepsilon_s$ with high probability, which is the condition we need in order to have Lemma \ref{lem:primal-opt} to hold.}

\begin{lemma}
Suppose \textsc{DualOpt} is executed with $\eta_1=1/8(B_f+B_AB_J\bar\lambda)$, $\eta_2=\sigma_\phi/B_A^2$, $\mu=\sigma_A^2/B_\phi$, $n_0= \frac{(1+B_A\bar\lambda)^4\kappa_4^4\ln^2 T}{B_\phi^2B_D^4\bar\rho^4}\vee \frac{N^2}{\underline\rho^4}$ with $\kappa_4 = 2\bar d(L_D\sqrt{N}\vee(B_f\sqrt{N}+B_r))\sqrt{N\ln(2NT)}$, $\kappa_5=\frac{32\kappa_3^2\kappa_6^2\bar\lambda B_A\ln^2 T}{\mu^2\eta_2\bar d\sqrt{N}}\vee \frac{16\kappa_1^4\kappa_6^2\bar\lambda^2B_D^4(1+\mu\eta_2)}{\mu^4\eta_2^2\bar d^2N}$ and $\kappa_6=\frac{2(B_\phi+\bar\lambda(B_A\bar d+\gamma_{\max})\sqrt{N})(1+\mu\eta_2)^{3/2}}{\mu\eta_2}$.
Then with probability $1-\tilde O(T^{-1})$ the following hold for all iterations $s$:
\begin{equation}
\lambda_s\in\Lambda\;\;\;\;\text{and}\;\;\;\; Q(\lambda_s)-Q(\lambda^*)\leq \bar\varepsilon_s,
\end{equation}
where $\lambda^*=\arg\min_{\lambda\in\mathbb R_+^M}Q(\lambda)\in\Lambda$ and $\bar\varepsilon_s=(1+\mu\eta_2)^{-s/2}\kappa_6$.
\label{lem:dual-opt}
\end{lemma}

\section{Regret analysis}\label{sec:regret}
The following theorem is our main result upper bounding the regret of Algorithm PD-NRM (Algorithm \ref{alg:grad-est}-\ref{alg:dual-opt}).
\begin{theorem}
The cumulative regret of Algorithm PD-NRM designed in Section~\ref{sec:algorithm} can be upper bounded by
$$
\left(\frac{\kappa_8\ln(T)}{\sqrt{\mu\eta_2\tilde\eta}} + \frac{2B_r(\kappa_3+L_D NB_A)\ln(T)}{\gamma_{\min}\sqrt{\mu\eta_2\tilde\eta}}\right)\times \sqrt{T},
$$
where $\mu=\sigma_A^2/B_\phi$, $\tilde\eta=\eta_1\sigma_D^2\sigma_\phi/2 = \sigma_D^2\sigma_\phi/16(B_f+B_AB_J\bar\lambda)$ and
\begin{align*}
\kappa_8 &:= B_f N + 1.2\kappa_1^2 (B_f+B_AB_J\bar\lambda) N + (2\kappa_3+\sqrt{\kappa_5})\bar\lambda\sqrt{N} \\
&= O(N^{3.25}\ln^{1.5}(NT)+N^{2.5}\ln^3(NT)).
\end{align*}
\label{thm:regret-upper-bound}
\end{theorem}

From Theorem \ref{thm:regret-upper-bound}, we can see that the regret can be bounded above by $\tilde O(N^{3.25}\sqrt{T})$, which slightly improves the best regret upper bound $\tilde O(N^{3.5}\sqrt{T})$ in the literature (see \citealt{miao2021network}). On top of the regret improvement, our approach is much more practical as it uses gradient descent instead of ellipsoid method which is known to be empirically inefficient. Moreover, since \citet{miao2021network} do the optimization on the demand rate domain, it involves translating back and forth between price and demand, which further makes the approach less effective. Mathematically, the overhead of translating back and forth between price and demand rate domains is reflected in an high-order regret term that is $o(\sqrt{T})$ asymptotically
but with a high-degree polynomial dependency on the problem dimension $N$ and $M$, making the algorithm far less practical when the $N$ or $M$ (number of product or resource types) is not very small.

Now let us provide a proof sketch of Theorem \ref{thm:regret-upper-bound}, whose details can be found in Section \ref{sec:proof_main_thm} in the appendix. Essentially, there are two parts of the regret we need to bound: the cumulative revenue deviation from the optimal price (i.e., $\sum_t (f(p^*)-f(p_t))$), and the cumulative violation of the primal feasible/resource inventory constraints. Let us fix an epoch $s$ and a loop $\tau$ inside the epoch (and we denote these time periods as $\mathcal{T}(s,\tau)$), by Lemma \ref{lem:dual-opt} and Lemma \ref{lem:primal-opt}, we have with high probability,
\begin{align*}
	\frac{f(p_{\tau})+f(\tilde p)}{2}\geq& f(p^*)-\tilde O(N^{3.25}n_\tau^{-1/2}),\\
	\frac{\langle a_j,D(p_{\tau})+D(\tilde p) \rangle}{2}\leq& \gamma_j+\tilde O(N^{2.5}n_\tau^{-1/2})\qquad\forall j\in[M],
\end{align*}
where $\tilde p$ is the price for demand balancing in \textsc{GradEst}. Combining with our choice of price perturbation factor $u$ in \textsc{GradEst}, we have
\begin{align*}
	\sum_{t\in\mathcal{T}(s,\tau)}f(p_t)\geq& n_\tau f(p^*)-\tilde O(N^{3.25}\sqrt{n_\tau}),\\
	\sum_{t\in\mathcal{T}(s,\tau)}\langle a_j,D(p_t) \rangle\leq& n_\tau\gamma_j+\tilde O(N^{2.5}\sqrt{n_\tau})\qquad\forall j\in[M].
\end{align*}
Since by algorithm design we have at most $O(\ln(T))$ loops in each epoch and $O(\ln(T))$ epochs, summing over all the loops and epochs, we have that 
\begin{align*}
\sum_{t=1}^{T}(f(p^*)-f(p_t))\leq&\tilde O(N^{3.25}\sqrt{T}),\\
\sum_{t=1}^{T}\langle a_j,D(p_t) \rangle\leq& \gamma_j T+\tilde O(N^{2.5}\sqrt{T})\qquad\forall j\in[M],
\end{align*}
which lead to the final regret we want.

\section{Numerical Experiment}\label{sec:num_exp}
In this section, we illustrate the numerical performance of our algorithm PD-NRM by comparing with two benchmark algorithms in the literature.
\begin{itemize}
	\item \textbf{BZ}: This is Algorithm 3 in \citet{besbes2012blind}, which is used as a benchmark in other NRM with demand learning literature as well (see e.g., \citealt{chen2019nonparametric,chen2019network}).
	\item \textbf{NSC}: This is the algorithm named \textit{Nonparametric Self-adjusting Control} (NSC) proposed in \citet{chen2019nonparametric}, which essentially studies the same problem as in this paper but with stronger conditions. 
\end{itemize}
Note that we did not study the performance of the algorithm in \citet{chen2019network} because their regret with respect to $T$ is $\tilde O(T^{4/5})$ which is significantly worse than $\tilde O(\sqrt{T})$. On the other hand, the regret of NSC is $\tilde O(T^{1/2+\epsilon})$ for arbitrarily small $\epsilon>0$ is nearly the same as $\tilde O(\sqrt{T})$. As a result, we choose NSC over the algorithm in \citet{chen2019network} as our main benchmark. 

As an illustrative example, we take $N=M=2$, $A=[1,1;0,2]$, $\gamma=[0.1,0.1]$. The demand function $D(p)$ is a logistic function defined as
\[
\begin{split}
D_1(p)=&\frac{\exp(0.4-1.5p_1)}{1+\exp(0.4-1.5p_1)+\exp(0.8-2p_2)},\\
D_2(p)=&\frac{\exp(0.8-2p_2)}{1+\exp(0.4-1.5p_1)+\exp(0.8-2p_2)},\\
\end{split}
\]
and the range of price is set as $[\underline p,\overline p]=[0.8,5]$. In this problem, we vary $T$ from $500$ to $10,000,000$ which means the initial inventory for each product is from $50$ to $1,000,000$. To obtain the performance of our algorithm, we run 50 independent instances and take the average. To evaluate the performance of each algorithm, we use the \textit{percentage loss} (of algorithm $\pi$) defined as $\text{Reg}(\pi)/\text{Rwd}(\pi^*)$. As we can see, this example is the same as the one used in \citet{chen2019nonparametric}. The only difference is the price sensitivity in \citealt{chen2019nonparametric} which is $[0.015,0.02]$ instead of $[1.5,2]$. However, with $[\underline p,\overline p]=[80,500]$ in the code of \citealt{chen2019nonparametric} (we thank George Chen for providing their code), it is indeed equivalent to our example through normalization. As a result, for two benchmark algorithms, we simply use the results obtained in \citet{chen2019nonparametric} directly. 

To run Algorithm PD-NRM, it is important to tune the learning parameters illustrated in Table \ref{tab:constants} for better empirical performance. While we recognize that optimal parameter tuning is non-trivial, we only did some ad-hoc manual tuning to obtain the parameters as in Table \ref{tab:input}.
\begin{table}[h!]
	\centering
	\caption{Input parameters of our algorithm xxx.}
	\label{tab:input}
	\begin{tabular}{c|c}
		\hline
		Constants& Value\\
		\hline
		$n_0$& $0.1\cdot N^4\ln^2(NT)$ \\
		$\kappa_1$& $n_0^{0.25}$ \\
		$\kappa_2$& $\kappa_5^{0.5}$\\
		$\kappa_3$& $8\cdot \kappa_1\cdot \sqrt{N^3\ln(2NT)}+12\cdot \kappa_1^2$\\ 
		$\kappa_5$& $2/3\cdot 10^{-8}\cdot(N^{5.5}\ln^3 (NT)+N^4\ln^6(NT))$\\
		$\kappa_6$& $\sqrt{N}$ \\
		$\eta_1$& 1 \\
		$\eta_2$& 1\\
		$\mu$& 1\\
		\hline
	\end{tabular}
\end{table}

\begin{figure}
	\centering
	\caption{Percentage loss of all algorithms}
	\label{fig:regret}
	\includegraphics[scale=0.7]{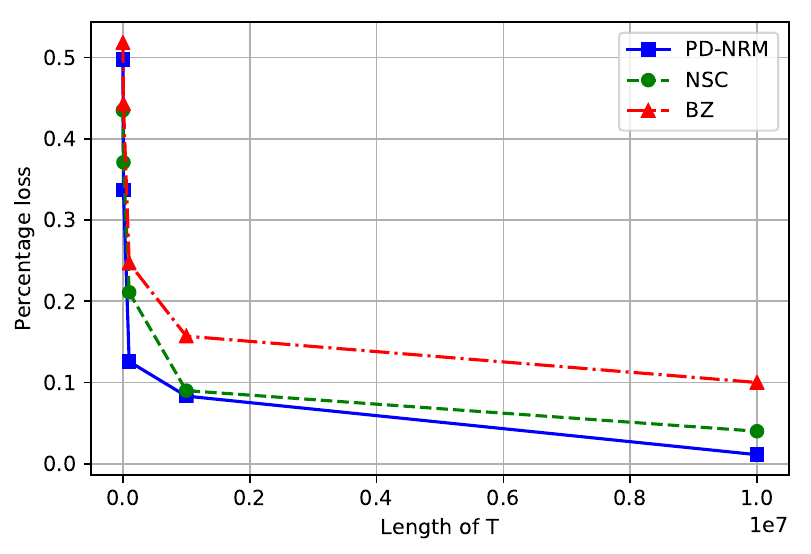}
\end{figure}

The performance of all algorithms interms of percentage loss is summarized in Figure \ref{fig:regret} and Table \ref{tab:regret}. From the results, we can see that our algorithm and NSC have better performance than BZ, while our algorithm also outperforms NSC in most of the cases. We shall note that the demand function we test is smooth so that NSC can be applied, and the dimension of our problem is only $N=2$. According to \citet{chen2019nonparametric}, it uses spline approximation to estimate the nonparametric demand function $D(\cdot)$, which requires a large number of base functions which is exponential in $N$, while our algorithm does not have this issue.

\begin{table}[h!]
	\centering
	\caption{Percentage loss of all algorithms}
	\label{tab:regret}
	\resizebox{0.8\textwidth}{!}{
		\renewcommand{\arraystretch}{1.5} 
		\begin{tabularx}{\textwidth}{|C|*{4}{>{\centering\arraybackslash}P{3.8cm}|}}
			\hline
			T=             & BZ (\%) & NSC (\%) & PD-NRM (\%) \\ \hline \hline
			500            & 52.6                       & 46.2                        & 53.0                            \\
			1,000          & 51.8                       & 43.5                        & 49.7                            \\
			2,000          & 50.1                       & 42.7                        & 44.6                            \\
			3,000          & 49.5                       & 43.8                        & 41.9                            \\
			4,000          & 48.3                       & 42.5                        & 37.0                            \\
			5,000          & 48.3                       & 40.3                        & 34.1                            \\
			6,000          & 47.5                       & 40.8                        & 34.7                            \\
			7,000          & 46.1                       & 40.4                        & 35.7                            \\
			8,000          & 46.3                       & 38.4                        & 34.6                            \\
			9,000          & 45.4                       & 38.8                        & 32.9                            \\
			$1\times 10^4$ & 44.3                       & 37.1                        & 33.7                            \\
			$1\times 10^5$ & 24.7                       & 21.1                        & 12.5                            \\
			$1\times 10^6$ & 15.7                       & 9.0                         & 8.3                             \\
			$1\times 10^7$ & 10.0                       & 4.0                         & 1.1                             \\ \hline
		\end{tabularx}
	}

\end{table}

\section{Conclusion}\label{sec:conclusion}
{
This paper proposes a practical algorithm named PD-NRM to solve the classical blind NRM problem, which is based on a primal-dual optimization scheme. In particular, we use Lagrangian dual variables $\lambda$ to relax the resource inventory constraints, and then search for an optimal price $p^*_{\lambda}$ corresponding to the Lagrangian with $\lambda$ fixed. The time horizon is divided into epochs with exponentially growing lengths for updating the dual variable; during each epoch, it is further divided into loops in a similar fashion for updating the price. More specifically, for both dual and primal optimization, we use some estimated (proximal) gradient descent for updating the decision variable, where the estimated gradients are outputs from the sub-routine \textsc{PrimalOpt} (see Algorithm \ref{alg:primal-opt}) and \textsc{GradEst} (see Algorithm \ref{alg:grad-est}) respectively. 

A key technical contribution of our paper is the demand balancing technique. In particular, we cannot simply apply the price we obtain from the estimated gradient descent in each loop because it may have significant violation of the resource inventory constraints due to the estimation error, which leads to sub-optimal regret in the end. To tackle that, we split each loop into two phases so that in the second phase, another price $\tilde p$ is computed to offset the violation of resource inventory constraint. With this technique, our algorithm PD-NRM is proved to have regret $\tilde O(N^{3.25}\sqrt{T})$ (see Theorem \ref{thm:regret-upper-bound}) which is, to the best of our knowledge, the best result in the literature. 

There are a few future research directions people may explore. The first direction is a technical issue we discussed under Assumption \ref{asmp:A}: how do we design algorithms which achieve similar performance when $N>M$ or $A$ does not have full row rank. Another research direction is to investigate how we can generalize the idea of demand balancing technique to other (potentially more general) online resource-constrained optimization. 

}

\bibliography{refs}
\bibliographystyle{apa-good}

\ECSwitch

\ECDisclaimer
%
%
%
\ECHead{Supplementary material: proofs of technical lemmas}

\section{Proof of results in Section~\ref{sec:fluid}}\label{sec:proof_sec3}

\subsection{Proof of Lemma \ref{lem:strong-convexity}}
When $d_\lambda^*$ lies in the interior of $D([\underline d,\overline d]^N)$, the $d_\lambda^*$ maximizing $\mH(\lambda,d)$ is unique because $\mH(\lambda,\cdot)$ is strongly concave.
The formula of $\nabla Q$ is then a direct consequence of the Danskin-Bertsekas theorem \citep{danskin2012theory,bertsekas1997nonlinear}.

We next derive the Hessian of $Q$. Because $d_\lambda^*$ lies in the interior of $D([\underline d,\overline d]^N)$, we have that $\partial_d \mH(\lambda,d_\lambda^*)=\nabla\phi(d_\lambda^*)-A^\top\lambda = 0$,
which holds for a neighborhood of $\lambda$ as well. Define $z:=A^\top\lambda = \nabla\phi(d_\lambda^*)$. By the chain rule and inverse function theorem, 
\begin{align*}
\nabla^2 Q(\lambda) &= -A\frac{\partial d_\lambda^*}{\partial\lambda} = -A\frac{\partial d_\lambda^*}{\partial z}\frac{\partial z}{\partial\lambda} =
-A\left[\frac{\partial z}{\partial d_\lambda^*}\right]^{-1}\frac{\partial z}{\partial\lambda} = -A\big[\nabla^2 \phi(d_\lambda^*)\big]^{-1}A^\top,
\end{align*} 
which is to be proved. 

\section{Proofs of results in Section~\ref{subsec:grad-est}}

\subsection{Proof of Lemma \ref{lem:est-D-JD}}
By vector Hoeffding's inequality \citep{tropp2012user} and the fact that $\|d\|_2\leq \bar d\sqrt{N}$ almost surely, it holds with probability $1-\tilde O(N^{-1}T^{-1})$ that 
\begin{equation}
\left\|\frac{1}{2N}\sum_{i=1}^N(d_{i,+} + d_{i,-}) - \frac{1}{2N}\sum_{i=1}^N(D(p_{i,+})+D(p_{i,-}))\right| \leq \bar d\sqrt{\frac{2N\ln(2NT)}{n}}.
\label{eq:proof-est-D-1}
\end{equation}
On the other hand, by Assumption \ref{asmp:D} the Jacobian of $D$ is Lipschitz continuous, and therefore by the mean-value theorem, for every $i\in[N]$, 
\begin{equation}
\|D(p_{i,\pm}) - (D(p) \pm u\partial_i D(p))\|_2 \leq \frac{1}{2}L_Du^2.
\label{eq:proof-est-D-2}
\end{equation}
Combining Eqs.~(\ref{eq:proof-est-D-1},\ref{eq:proof-est-D-2}) and noting that the $\pm u\partial_i D(p)$ terms in Eq.~(\ref{eq:proof-est-D-2}) are canceled out, we prove the first property of 
Lemma \ref{lem:est-D-JD}.

Next fix $i\in[N]$. Dividing both sides of Eq.~(\ref{eq:proof-est-D-2}) by $2u$, we obtain
\begin{equation}
\left\|\frac{D(p_{i,+})-D(p_{i,-})}{2u} - \partial_i D(p)\right\|_2\leq \frac{1}{2}L_D u.
\label{eq:proof-est-D-3}
\end{equation}
Similarly, dividing both sides of Eq.~(\ref{eq:proof-est-D-1}) by $2u$ (for a particular $i\in[N]$) we obtain
\begin{equation}
\left\|\frac{d_{i,+}-d_{i,-}}{2u} - \frac{D(p_{i,+})-D(p_{i,-})}{2u}\right\|_2 \leq \frac{2\bar d N}{u}\sqrt{\frac{\ln(2NT)}{n}}.
\label{eq:proof-est-D-4}
\end{equation}
Combining Eqs.~(\ref{eq:proof-est-D-3},\ref{eq:proof-est-D-4}) we prove the second property of Lemma \ref{lem:est-D-JD}. 

Finally we prove the third property. Fix an arbitrary $i\in[N]$. By the mean-value theorem and Assumption \ref{asmp:f-phi}, 
\begin{equation}
\big|f(p_{i,\pm}) - (f(p) \pm u\partial_i f(p))\big| \leq \frac{1}{2}B_f u^2. 
\label{eq:proof-est-D-5}
\end{equation}
Dividing both sides of Eq.~(\ref{eq:proof-est-D-5}) and do a differentiation at $p_{i,+}$ minus $p_{i,-}$, we obtain
\begin{equation}
\left|\frac{f(p_{i,+})-f(p_{i,-})}{2u} - \partial_i f(p)\right| \leq \frac{B_f u}{2}.
\label{eq:proof-est-D-6}
\end{equation}
On the other hand, note that $f(p_{i,\pm}) = \mathbb E[\langle p_{i,\pm},d_{i,\pm}\rangle]$, and that $0\leq \langle p_{i,\pm},d_{i,\pm}\rangle\leq B_r$ almost surely. By Hoeffding's inequality, it holds with probability $1-\tilde O(N^{-1}T^{-1})$ that
$$
\big| \langle p_{i,\pm}, d_{i,\pm}\rangle - f(p_{i,\pm})\big| \leq B_r\sqrt{\frac{2N\ln(2NT)}{n}}.
$$
Dividing both sides of the above inequality by $2u$ and do a differentiation at $p_{i,+}$ minus $p_{i,-}$, we obtain
\begin{equation}
\left|\frac{\langle p_{i,+},d_{i,+}\rangle - \langle p_{i,-},d_{i,-}\rangle}{2u} - \frac{f(p_{i,+})-f(p_{i,-})}{2u}\right|\leq \frac{B_r}{u}\sqrt{\frac{2N\ln(2NT)}{n}}.
\label{eq:proof-est-D-7}
\end{equation}
Combining Eqs.~(\ref{eq:proof-est-D-6},\ref{eq:proof-est-D-7}) and extend to all product types $i\in[N]$ we complete the proof of the third property.

\subsection{Proof of Lemma \ref{lem:feasibility}.}

Let $\tilde p$ be any price vector feasible to Line \ref{line:tilde-p} and fix $j\in[M]$.
By the mean-value theorem, there exists $\xi\in[0,1]$ such that $D(\tilde p)=D(p)+J_D(\xi p+(1-\xi)\tilde p)(\tilde p-p)$. Using the Lipschitz continuity of $J_D$ in Assumption \ref{asmp:D}, we obtain
\begin{align}
\| D(\tilde p)- (D(p)+J_D(p)(\tilde p-p))\|_2 &\leq \|J_D(\xi p+(1-\xi)\tilde p)-J_D(p)\|_2\|\tilde p-p\|_2\nonumber\\
& \leq L_D\|\tilde p-p\|_2^2 \leq L_D\kappa_1^2n^{-1/2}.
\label{eq:proof-feasibility-1}
\end{align}
On the other hand, invoking Lemma \ref{lem:est-D-JD} and noting that $u=\sqrt{N}/\sqrt[4]{n}$ with $p$ satisfying the conditions in Lemma \ref{lem:feasibility}, we have that 
\begin{align}
\|\hat D(p)-D(p)\|_2 &\leq 2\bar d\sqrt{N}(\sqrt{\ln NT}+L_D\sqrt{N})n^{-1/2};\label{eq:proof-feasibility-2}\\
\|\hat J_D(p)-J_D(p)\|_\op &\leq (2\bar d\sqrt{\ln(2NT)}+L_D) Nn^{-1/4}.\label{eq:proof-feasibility-3}
\end{align}
Combine Eqs.~(\ref{eq:proof-feasibility-1},\ref{eq:proof-feasibility-2},\ref{eq:proof-feasibility-3}) and note that $\|\tilde p-p\|_2\leq \kappa_1\sqrt{N} n^{-1/4}$. We have
\begin{align}
\bigg\|\hat D(p) &+ \frac{1}{2}\hat J_D(p)(\tilde p-p) - \frac{D(p)+D(\tilde p)}{2}\bigg\|_2\leq 2\bar d\sqrt{N}(\sqrt{\ln NT}+L_D\sqrt{N})n^{-1/2}\nonumber\\
&+ (2\bar d\sqrt{\ln(2NT)}+L_D) Nn^{-1/4}\times \kappa_1\sqrt{N} n^{-1/4} + L_D\kappa_1^2n^{-1/2},
\end{align}
and therefore 
\begin{align}
\bigg|\left\langle a_j,\hat D(p) + \frac{1}{2}\hat J_D(p)(\tilde p-p) - \frac{D(p)+D(\tilde p)}{2}\right\rangle\bigg| &\leq B_A\bigg\|\hat D(p) + \frac{1}{2}\hat J_D(p)(\tilde p-p) - \frac{D(p)+D(\tilde p)}{2}\bigg\|_2 \nonumber\\
&\leq (\kappa_3-2B_AL_D\kappa_1^2)/\sqrt{n},\;\;\;\;\;\;\forall j\in[M], \label{eq:proof-feasibility-3half}
\end{align}
because $\|a_j\|_2\leq \|A\|_\op\leq B_A$ thanks to Assumption \ref{asmp:A}. The first property in Lemma \ref{lem:feasibility} is then proved by applying the Cauchy-Schwarz inequality.

We next prove the second property. Construct $\tilde p = 2p^*-p$, which satisfies the first constraint on Line \ref{line:tilde-p},
 $\|\tilde p-p\|_2\leq \kappa_1 n^{-1/4}$, by the condition that $\|p-p^*\|_2\leq 0.5\kappa_1 n^{-1/4}$.
Applying the mean-value theorem at $p^*$ and using the Lipschitz continuity of $J_D$ thanks to Assumption \ref{asmp:D}, it holds that
\begin{align}
\|D(\tilde p)-(D(p^*)+J_D(p^*)(\tilde p-p^*)\|_2 &\leq L_D\kappa_1^2 n^{-1/2};\label{eq:proof-feasibility-4}\\
\|D(p)-(D(p^*)+J_D(p^*)(p-p^*)\|_2 &\leq L_D\kappa_1^2 n^{-1/2}.\label{eq:proof-feasibility-5}
\end{align}
With $\tilde p+p=2p^*$, combining Eqs.~(\ref{eq:proof-feasibility-4},\ref{eq:proof-feasibility-5}) we obtain
\begin{equation}
\left\|\frac{D(p)+D(\tilde p)}{2} - D(p^*)\right\|_2 \leq 2L_D\kappa_1^2 n^{-1/2}.
\label{eq:proof-feasibility-6}
\end{equation}
This together with Eq.~(\ref{eq:proof-feasibility-3half}) and the fact that $AD(p^*)\leq\gamma$ since $p^*$ is primal feasible, establishes that the second constraint on Line \ref{line:tilde-p} is satisfied.

We next establish the third constraint. If $\lambda_j^*>0$, then by complementary slackness $\langle a_j,D(p^*)\rangle = \gamma_j$ and therefore the third constraint is satisfied,
by a simple symmetric argument. Now consider any $j\in[M]$ with $\lambda_j^*=0$.
Because $Q$ is convex, we have that $\langle\nabla Q(\lambda^*),\lambda-\lambda^*\rangle \leq Q(\lambda) - Q(\lambda^*)\leq \kappa_2/\sqrt{n}$.
On the other hand, because $\nabla Q(\lambda^*)=\gamma - AD(p^*)\geq 0$ since $AD(p^*)\leq \gamma$ thanks to primal feasibility, and $\partial_k Q(\lambda^*)=0$ whenever $\lambda_k^*>0$, we know that $\partial_k Q(\lambda^*)(\lambda_k-\lambda_k^*)\geq 0$ for all $k$. Hence, $\partial_j Q(\lambda^*)(\lambda_j-\lambda_j^*)=\partial_j Q(\lambda^*)\lambda_j\leq \kappa_2/\sqrt{n}$, which implies
\begin{equation}
\partial_j Q(\lambda^*) = \gamma_j - \langle a_j, D(p^*)\rangle \leq \frac{\kappa_2}{\lambda_j\sqrt{n}}\leq \frac{\kappa_2}{(1\wedge \lambda_j)\sqrt{n}}.
\label{eq:proof-feasibility-7}
\end{equation}
Combining Eq.~(\ref{eq:proof-feasibility-7}) with Eqs.~(\ref{eq:proof-feasibility-6},\ref{eq:proof-feasibility-3half}) we established that the third constraint on Line \ref{line:tilde-p} is feasible too.

\section{Proofs of results in Section~\ref{subsec:primal-opt}}

\subsection{Proof of Lemma \ref{lem:pl}}
Let $d=D(p)$, $d_\lambda^*=D(p_\lambda^*)$ and recall the definition that $\mH(\lambda,d)=\phi(d)-\langle \lambda, Ad-\gamma\rangle = \mL(\lambda,p)$.
By the chain rule, we have that
\begin{equation}
\nabla_p\mL(\lambda,p) = J_D(p)^\top\nabla_d\mH(\lambda,d).
\label{eq:proof-pl-1}
\end{equation}
Hence,
\begin{align}
\frac{1}{2}\|\nabla_p\mL(\lambda,p)\|_2^2 = \frac{1}{2}\nabla_d\mH(\lambda,d)^\top (J_D(p)J_D(p)^\top) \nabla_d\mH(\lambda,d) \geq \frac{\sigma_D^2}{2}\|\nabla_d\mH(\lambda,d)\|_2^2,
\label{eq:proof-pl-1half}
\end{align}
where the last inequality holds because $J_D(p)$ is square and $\sigma_{\min}(J_D(p))\geq \sigma_D$ thanks to Assumption \ref{asmp:D}.
On the other hand, note that $\mH(\lambda,\cdot)$ is smooth and $\sigma_\phi$-strongly concave in $d$, thanks to Assumption \ref{asmp:f-phi}. Also, $d_\lambda^*$ is an interior maximizer thanks to Assumption \ref{asmp:Lambda}, which implies $\nabla_d\mH(\lambda,d_\lambda^*)=0$. Subsequently, for some $\xi\in[0,1]$ it holds that
\begin{align}
\frac{1}{2}\|\nabla_d\mH(\lambda,d)\|_2^2 &= \frac{1}{2}\|\nabla_d^2 \mH(\lambda,\xi d+(1-\xi)d_\lambda^*)(d-d_\lambda^*)\|_2^2 \geq \frac{\sigma_\phi^2}{2}\|d-d_\lambda^*\|_2^2\nonumber\\
& \geq \sigma_\phi[\mH(\lambda,d_\lambda^*)-\mH(\lambda,d)] = \sigma_\phi[\mL(\lambda,p_\lambda^*)-\mL(\lambda,p)].
\label{eq:proof-pl-2}
\end{align}
Combining Eqs.~(\ref{eq:proof-pl-1half},\ref{eq:proof-pl-2}) we complete the proof of the first inequality.

To prove the second inequality, note that, 
\begin{align}
\mH(\lambda,d) &\leq \mH(\lambda,d_\lambda^*) - \frac{\sigma_\phi}{2}\|d-d_\lambda^*\|_2^2 = \mH(\lambda,d_\lambda^*) - \frac{\sigma_\phi}{2}\|D(p)-D(p_\lambda^*)\|_2^2\nonumber\\
&= \mH(\lambda,d_\lambda^*) - \frac{\sigma_\phi}{2}\|J_D(\xi p+(1-\xi)p_\lambda^*)(p-p_\lambda^*)\|_2^2\label{eq:proof-pl-3}\\
&\leq \mH(\lambda,d_\lambda^*) - \frac{\sigma_\phi\sigma_D^2}{2}\|p-p_\lambda^*\|_2^2,\label{eq:proof-pl-4}
\end{align}
where Eq.~(\ref{eq:proof-pl-3}) holds for some $\xi\in[0,1]$ thanks to the mean value theorem, and Eq.~(\ref{eq:proof-pl-4}) holds because $\sigma_{\min}(J_D(p))\geq\sigma_D$ for all $p$,
thanks to Assumption \ref{asmp:D}. This proves the second inequality in Lemma \ref{lem:pl}.

\subsection{Proof of Lemma \ref{lem:primal-opt}}
We first prove the upper bound on $\|p_\tau-p_\lambda^*\|_2$. Let $w_\tau := \hat\nabla_p\mL(\lambda,p_\tau)-\nabla_p\mL(\lambda,p_\tau)$ be the error of gradient estimation at iteration $\tau$,
and $\varepsilon_\tau := \|w_\tau\|_2$ be its magnitude. 
Note that $\mL(\lambda,\cdot)$ has Lipschitz continuous gradients, or more specifically for any $p,p'$, 
\begin{align}
\|\nabla_p\mL(\lambda,p)-\nabla_p\mL(\lambda,p')\|_2 \leq \|\nabla f(p)-\nabla f'(p)\|_2 + B_A\bar\lambda \|J_D(p)-J_D(p')\|_\op\leq (B_f+B_AB_J\bar\lambda)\|p-p'\|_2,
\label{eq:proof-primal-opt-0}
\end{align}
where the last inequality holds thanks to Assumptions \ref{asmp:f-phi}, \ref{asmp:D}, and the definition that $\bar\lambda=\sup\{\|\lambda\|_2: \lambda\in\Lambda\}$.
Subsequently, by the mean-value theorem it holds that, for some $\xi\in[0,1]$, 
\begin{align}
\mL(\lambda,p_{\tau+1}) &= \mL(\lambda,p_\tau) + \langle\nabla_p\mL(\lambda,\xi p_\tau+(1-\xi)p_{\tau+1}), p_{\tau+1}-p_\tau\rangle\nonumber\\
&\geq \mL(\lambda,p_\tau) + \langle\nabla_p\mL(\lambda,p_\tau), p_{\tau+1}-p_\tau\rangle - (B_f+B_AB_J\bar\lambda)\|p_{\tau+1}-p_\tau\|_2^2\label{eq:proof-primal-opt-1}.
\end{align}
Note that, with $p_{\tau+1}-p_\tau = \eta\hat\nabla_p\mL(\lambda,p_\tau) = \eta(\nabla_p\mL(\lambda,p_\tau)+w_\tau)$. We have that
\begin{align}
\langle\nabla_p\mL(\lambda,p_\tau), p_{\tau+1}-p_\tau\rangle &= \eta\|\nabla_p\mL(\lambda,p_\tau)\|_2^2  + \eta\langle\nabla_p\mL(\lambda,p_\tau), w_\tau\rangle\nonumber\\
&\geq \eta\|\nabla_p\mL(\lambda,p_\tau)\|_2^2  - \frac{\eta}{2}\|\nabla_p\mL(\lambda,p_\tau)\|_2^2 - \frac{\eta}{2}\|w_\tau\|_2^2\nonumber\\
&\geq \frac{\eta}{2}\|\nabla_p\mL(\lambda,p_\tau)\|_2^2 - \frac{\eta}{2}\varepsilon_\tau^2;\label{eq:proof-primal-opt-1first}\\
\|p_{\tau+1}-p_\tau\|_2^2 &= \eta^2\|\nabla_p\mL(\lambda,p_\tau)+w_\tau\|_2^2 \leq 2\eta^2\|\nabla_p\mL(\lambda,p_\tau)\|_2^2 + 2\eta^2\varepsilon_\tau^2.\label{eq:proof-primal-opt-1second}
\end{align}
Incorporating Eqs.~(\ref{eq:proof-primal-opt-1first},\ref{eq:proof-primal-opt-1second}) into Eq.~(\ref{eq:proof-primal-opt-1}), we obtain 
\begin{align}
\mL(\lambda,p_{\tau+1}) &\geq \mL(\lambda,p_\tau) + \frac{\eta}{2}\|\nabla_p\mL(\lambda,p_\tau)\|_2^2 - \frac{\eta}{2}\varepsilon_\tau^2 - 2\eta^2(B_f+B_AB_J\bar\lambda)(\|\nabla_p\mL(\lambda,p_\tau)\|_2^2+\varepsilon_\tau^2).
\label{eq:proof-primal-opt-1half}
\end{align}
With $\eta=1/8(B_f+B_AB_J\bar\lambda)$, Eq.~(\ref{eq:proof-primal-opt-1}) can be simplified to
\begin{equation}
\mL(\lambda,p_{\tau+1}) \geq  \mL(\lambda,p_\tau) + \frac{\eta}{4}\|\nabla_p\mL(\lambda,p_\tau)\|_2^2 - \eta\varepsilon_\tau^2.
\label{eq:proof-primal-opt-2}
\end{equation}
Using the PL inequality established in Lemma \ref{lem:pl}, Eq.~(\ref{eq:proof-primal-opt-2}) is further simplified to
\begin{equation}
\mL(\lambda,p_{\tau+1}) \geq   \mL(\lambda,p_\tau)  + \frac{\eta\sigma_D^2\sigma_\phi}{2}[\mL(\lambda,p_\lambda^*)-\mL(\lambda,p_\tau)] - \eta\varepsilon_\tau^2.
\label{eq:proof-primal-opt-3}
\end{equation}
Let $\delta_\tau := \mL(\lambda,p_\lambda^*)-\mL(\lambda,p_\tau)$ and $\tilde\eta := \eta\sigma_D^2\sigma_\phi/2$. Re-arranging terms in Eq.~(\ref{eq:proof-primal-opt-3}), we obtain
\begin{equation}
\delta_{\tau+1} \leq (1-\tilde\eta)\delta_\tau + \eta\varepsilon_\tau^2.
\label{eq:proof-primal-opt-4}
\end{equation}
Expanding Eq.~(\ref{eq:proof-primal-opt-4}) all the way to $\tau=0$, we obtain for every $\tau$ that
\begin{equation}
\delta_\tau \leq (1-\tilde\eta)^\tau \delta_0 + \eta \sum_{k=0}^{\tau-1}(1-\tilde\eta)^{\tau-k+1}\varepsilon_k^2.
\label{eq:proof-primal-opt-5}
\end{equation}

Note that for $p_0\in P$, $\|p_0-p_\lambda^*\|_2\leq\bar\rho$ thanks to Assumption \ref{asmp:Lambda}. Hence, $\|d_0-d_\lambda^*\|_2\leq B_D\bar\rho$, where $d_0=D(p_0)$.
Note also that $\nabla_d\mH(\lambda,d_\lambda^*)=0$ because $d_\lambda^*$ is an interior maximizer thanks to Assumption \ref{asmp:Lambda}, 
and $-\nabla_d^2\mH(\lambda,d) = -\nabla^2\phi(d)\preceq B_\phi I_{N\times N}$ thanks to Assumption \ref{asmp:f-phi}, we have $\mH(\lambda,d_\lambda^*)-\mH(\lambda,d)\leq \frac{B_\phi}{2}B_D^2\bar\rho^2$. This means that we can simply take $\delta_0=\frac{B_\phi}{2}B_D^2\bar\rho^2$ as an upper bound of $\mL(\lambda,p_\lambda^*)-\mL(\lambda,p_0)$ without affecting our analysis.

With the choice $n_k=\lceil (1-\tilde\eta)^{-2k}n_0\rceil$ for sufficiently large $n_0$ (we will calculate precisely the value of $n_0$ later) so that $u=\sqrt{N}/\sqrt[4]{n_k}$ at iteration $k$ in the \textsc{GradEst} sub-routine, Lemma \ref{lem:est-D-JD} implies that with probability $1-\tilde O(T^{-1})$, 
\begin{align*}
\|\hat \nabla f(p_\tau)-\nabla f(p_\tau)\|_2 &\leq \frac{\kappa_4}{\sqrt[4]{n_\tau}} \leq \frac{\kappa_4(1-\tilde\eta)^{\tau/2}}{\sqrt[4]{n_0}};\\
\|\hat J_D(p_\tau)-J_D(p_\tau)\|_\op &\leq  \frac{\kappa_4}{\sqrt[4]{n_\tau}} \leq \frac{\kappa_4(1-\tilde\eta)^{\tau/2}}{\sqrt[4]{n_0}},
\end{align*}
where $\kappa_4 = 2\bar d(L_D\sqrt{N}\vee(B_f\sqrt{N}+B_r))\sqrt{N\ln(2NT)}$. Subsequently,
$$
\varepsilon_\tau = \|\hat\nabla_p\mL(\lambda,p_\tau)-\nabla_p\mL(\lambda,p_\tau)\|_2 \leq (1+B_A\bar\lambda)\kappa_4 (1-\tilde\eta)^{\tau/2} n_0^{-1/4}.
$$
Subsequently, the above inequality together with Eq.~(\ref{eq:proof-primal-opt-5}) yields
\begin{equation}
\delta_\tau \leq (1-\tilde\eta)^\tau\left(\delta_0 + \tau \eta (1+B_A\bar\lambda)^2\kappa_4^2n_0^{-1/2}\right).
\label{eq:proof-primal-opt-6}
\end{equation}
Note that the total number of iterations $\tau_0$ is upper bounded by $\tau_0\leq \ln(T)/(-2\ln(1-\eta))$. With the choice of $n_0$ such that 
\begin{equation}
n_0\geq \frac{(1+B_A\bar\lambda)^4\kappa_4^4\ln^2 T}{B_\phi^2B_D^4\bar\rho^4},
\label{eq:proof-n0-ub}
\end{equation}
and noting that $\ln^2(1-\eta)\geq \eta^2$ for all $\eta\in(0,1)$, and that $\delta_0 = \frac{B_\phi}{2}B_D^2\bar\rho^2$, the second term in Eq.~(\ref{eq:proof-primal-opt-6}) is upper bounded by $\delta_0$. Consequently, 
\begin{equation}
\mL(\lambda,p_\lambda^*)-\mL(\lambda,p_\tau) = \delta_\tau \leq 2(1-\tilde\eta)^\tau \delta_0 \leq B_\phi B_D^2\bar\rho^2(1-\tilde\eta)^\tau \leq B_\phi B_D^2\bar\rho^2\sqrt{\frac{n_0}{n_\tau}},
\label{eq:proof-primal-opt-7}
\end{equation}
where the last inequality is due to the definition of $n_\tau$. This proves the first property in Lemma \ref{lem:primal-opt}. Invoking the quadratic growth property in Lemma \ref{lem:pl}, the above inequality is further simplified to
\begin{align}
\|p_\lambda^*-p_\tau\|_2 &\leq \sqrt{\frac{2(\mL(\lambda,p_\lambda^*)-\mL(\lambda,p_\tau))}{\sigma_\phi\sigma_D^2}} \leq \sqrt{\frac{2B_\phi B_D^2\bar\rho^2}{\sigma_\phi\sigma_D^2}}\sqrt[4]{\frac{n_0}{n_\tau}},
\label{eq:proof-primal-opt-7half}
\end{align}
which proves the second property in Lemma \ref{lem:primal-opt}.

To prove the third property, note that Eqs.~(\ref{eq:proof-primal-opt-6},\ref{eq:proof-n0-ub}) together imply that $\mL(\lambda,p_\lambda^*)-\mL(\lambda,p_\tau)\leq 2\delta_0 \leq B_\phi B_D^2\bar\rho^2$ for all $\tau$. Hence, by Lemma \ref{lem:pl} it holds that 
$$
\|p_\lambda^*-p_\tau\|_2 \leq \sqrt{\frac{2B_\phi B_D^2\bar\rho^2}{\sigma_\phi\sigma_D^2}} = \frac{B_D}{\sigma_D}\sqrt{\frac{2B_\phi}{\sigma_\phi}}\bar\rho = \chi\bar\rho,
$$ 
where $\chi$ is the constant specified in Assumption \ref{asmp:Lambda}.  The third property of Lemma \ref{lem:primal-opt} is then implied by the fourth item in Assumption \ref{asmp:Lambda}.

With the choice of
$$
n_0\geq \frac{N^2}{\underline\rho^4}
$$
and the third property, it holds that $u=\sqrt{N}/\sqrt[4]{n}$ in the invocation of \textsc{GradEst}.

We next focus on the last two properties in Lemma \ref{lem:primal-opt}. With the choices that $\kappa_1=\sqrt{\frac{8B_\phi B_D^2\bar\rho^2}{\sigma_\phi\sigma_D^2}}\sqrt[4]{n_0}$ and $\kappa_2=\sqrt{\kappa_5}$,
together with the stopping rule of $n_\tau>\kappa_5/\bar\varepsilon^2$ for some $\kappa_5\geq 1$, it is easy to check through Eq.~(\ref{eq:proof-primal-opt-7half}) that $\|p_\lambda^*-p_\tau\|_2\leq \kappa_1/2\sqrt[4]{n_\tau}$, and that
$Q(\lambda^*)-Q(\lambda)\leq \kappa_2/\sqrt{n_\tau}$. By Lemma \ref{lem:feasibility}, this implies that, with probability $1-\tilde O(T^{-1})$, the problem on Line \ref{line:tilde-p} in the
\textsc{GradEst} sub-routine is feasible, and therefore with high probability $\tilde p\neq p$ satisfies all properties listed in Lemma \ref{lem:feasibility}. More specifically, 
$$
\frac{\langle a_j, D(p_\tau)+D(\tilde p_\tau)\rangle}{2}\leq \gamma_j + \frac{\kappa_3}{\sqrt{n_\tau}},
$$
which establishes the last property in Lemma \ref{lem:primal-opt}. Finally, to prove the second to last inequality, note that for any $p$, $f(p)=\mL(\lambda,p) +\langle \lambda, AD(p)-\gamma\rangle$. 
Subsequently, 
\begin{align}
&\frac{f(p_\tau)+f(\tilde p_\tau)}{2} = \frac{\mL(\lambda,p_\tau)+\mL(\lambda,\tilde p_\tau)+\langle \lambda , AD(p_\tau)+AD(\tilde p_\tau)-2\gamma\rangle}{2} \nonumber\\
&= Q(\lambda) - \frac{\mL(\lambda,p_\lambda^*)-\mL(\lambda,p_\tau)}{2} - \frac{\mL(\lambda,p_\lambda^*)-\mL(\lambda,\tilde p_\tau)}{2} + \left\langle\lambda, A\frac{D(p_\tau)+D(\tilde p_\tau)}{2}-\gamma\right\rangle\nonumber\\
&\geq f(p^*) - \frac{\mL(\lambda,p_\lambda^*)-\mL(\lambda,p_\tau)}{2} - \frac{\mL(\lambda,p_\lambda^*)-\mL(\lambda,\tilde p_\tau)}{2} + \left\langle\lambda, A\frac{D(p_\tau)+D(\tilde p_\tau)}{2}-\gamma\right\rangle\label{eq:proof-primal-opt-8}\\
&\geq f(p^*) - (B_f+B_AB_J\bar\lambda)\left(\frac{\|p_\lambda^*-p_\tau\|_2^2}{2} + \frac{\|p_\lambda^*-\tilde p_\tau\|_2^2}{2}\right) + \left\langle\lambda, A\frac{D(p_\tau)+D(\tilde p_\tau)}{2}-\gamma\right\rangle
\label{eq:proof-primal-opt-9}\\
&\geq f(p^*) - (B_f+B_AB_J\bar\lambda)\frac{1.2\kappa_1^2 N}{\sqrt{n_\tau}} + \left\langle\lambda, A\frac{D(p_\tau)+D(\tilde p_\tau)}{2}-\gamma\right\rangle\label{eq:proof-primal-opt-10}\\
&\geq  f(p^*) - (B_f+B_AB_J\bar\lambda)\frac{1.2\kappa_1^2 N}{\sqrt{n_\tau}}  - \sum_{j=1}^m \lambda_j \left(\frac{\kappa_2}{(1\wedge\lambda_j)\sqrt{n_\tau}} +\frac{2\kappa_3}{\sqrt{n_\tau}}\right)\label{eq:proof-primal-opt-11}\\
&\geq f(p^*) - (B_f+B_AB_J\bar\lambda)\frac{1.2\kappa_1^2 N}{\sqrt{n_\tau}} - \frac{(2\kappa_3+\sqrt{\kappa_5})\bar\lambda\sqrt{N}}{\sqrt{n_\tau}}.
\end{align}
Here, Eq.~(\ref{eq:proof-primal-opt-8}) holds by the definition that $\lambda^*=\arg\min_{\lambda\in\mathbb R_+^M}Q(\lambda)$, and that $Q(\lambda^*)=f(p^*)$ thanks to strong duality as established in Lemma \ref{lem:strong-duality}; 
Eq.~(\ref{eq:proof-primal-opt-9}) holds because $\mL(\lambda,\cdot)$ is smooth and $p_\lambda^*$ is its interior maximizer, yielding $\nabla_p\mL(\lambda,p_\lambda^*)=0$ and therefore with Eq.~(\ref{eq:proof-primal-opt-0}) and the mean-value theorem, 
\begin{align*}
\mL(\lambda,p_\lambda^*)-\mL(\lambda,p) \leq (B_f+B_AB_J\bar\lambda)\|p-p_\lambda^*\|_2^2,\;\;\;\;\;\;\forall p;
\end{align*}
Eq.~(\ref{eq:proof-primal-opt-10}) holds because by the choice of parameter $\kappa_1$, $\|p_\lambda^*-p_\tau\|_2\leq \kappa_1/(2\sqrt[4]{n_\tau})$ with high probability,
and $\|\tilde p_\tau-p_\tau\|_2\leq \sqrt{N}\|\tilde p_\tau-p_\tau\|_{\infty} \leq \kappa_1\sqrt{N}/\sqrt[4]{n_\tau}$ thanks to the constraint on Line \ref{line:tilde-p} of Algorithm \ref{alg:grad-est};
Eq.~(\ref{eq:proof-primal-opt-11}) holds by invoking the first property in Lemma \ref{lem:feasibility}, and noting that $\kappa_2=\sqrt{\kappa_5}$. This proves the fourth property in Lemma \ref{lem:primal-opt}.

\subsection{Proof of Lemma \ref{lem:dual-opt}}

To prove the lemma, we first present a technical result upper bounding a properly average sub-optimality gaps of proximal gradient descent when gradients contain errors.
\begin{lemma}
For each iteration $s$ in Algorithm \ref{alg:dual-opt}, let $e_s := \hat\nabla Q(\lambda_s)-\nabla Q(\lambda_s)$. Suppose also that $\mu=\sigma_A/B_\phi^2$ and $\eta_2=\sigma_\phi^2/B_A$. Then for any $m\geq 1$, 
$$
\frac{\sum_{s=1}^s c_{s-1}[Q(\lambda_s)-Q(\lambda^*)]}{\sum_{s=1}^m c_{s-1}} \leq \frac{(1+\mu\eta_2)^{-m+1}}{2\eta_2}\|\lambda^*-\lambda_0\|_2^2 + 2(1+\mu\eta_2)^{-m+1}\left(\sum_{s=0}^{m-1}\sqrt{\eta_2}(1+\mu\eta_2)^{s/2}\|e_s\|_2\right)^2,
$$
where $c_s := (1+\mu\eta_2)^{s}=(1+\sigma_\phi^2\sigma_A/B_\phi^2B_A)^s$.
\label{lem:inexact-md}
\end{lemma}

At a higher level, Lemma \ref{lem:inexact-md} is similar to \citep[Proposition 3]{schmidt2011convergence}, analyzing proximal gradient descent with inexact gradients.
However, Proposition 3 of \cite{schmidt2011convergence} only upper bounds the distance between $\lambda_s$ and $\lambda^*$ in $\ell_2$-norm, which is insufficient for our purpose
because our $\lambda^*$ may well lie on the boundary of $\Lambda$ (e.g., in the form of non-binding inventory constraints for certain resource types).
Instead, here in Lemma \ref{lem:inexact-md} we directly prove upper bounds on the sub-optimality gap, via a proper averaging argument.

We now carry out the proof by an inductive argument on the iteration $s$. For $s=0$, the conclusion that $\lambda_0\in\Lambda$ and $Q(\lambda_0)-Q(\lambda^*)\leq \kappa_6$ clearly holds because 
$$
\kappa_6\geq B_\phi+\bar\lambda(B_A\bar d+\gamma_{\max})\sqrt{N},
$$ 
because $Q(\lambda^*)=\phi(d^*)>0$ and $Q(\lambda)\leq B_\phi+\bar\lambda(B_A\bar d+\gamma_{\max})\sqrt{N}$ for all $\lambda\in\Lambda$. We next prove the desired properties for iteration $s+1$, assuming they hold for iteration $s$.

Note that for iteration $s+1$ in \textsc{Dual-Opt}, the last iteration $\tau_0$ in \textsc{Primal-Opt} satisfies $n_{\tau_0}\geq \kappa_5/2\bar\varepsilon_{s+1}^2$, where $\bar\varepsilon_{s+1}=(1+\mu\eta_2)^{-(s+1)/2}\kappa_6$.
By inductive hypothesis, all conditions in Lemma \ref{lem:primal-opt} are met and therefore with probability $1-\tilde O(T^{-1})$ it holds that
\begin{equation}
\|\hat p(\lambda_s)-p_{\lambda_s}^*\|_2 \leq \frac{\kappa_1}{2\sqrt[4]{n_{\tau_0}}}\leq \frac{\kappa_1}{\sqrt[4]{\kappa_5}}\sqrt{\bar\varepsilon_s},
\label{eq:proof-dual-opt-1}
\end{equation}
where $p_{\lambda_s}^*=\arg\max_{p} \mL(\lambda,p)$.
By Lemma \ref{lem:est-D-JD}, it also holds with probability $1-\tilde O(T^{-1})$ that
\begin{equation}
\|\hat D(\hat p(\lambda_s)) - D(\hat p(\lambda_s))\|_2 \leq \frac{2\bar dN(\sqrt{\ln(2NT)}+L_D\sqrt{N})}{\sqrt{n_{\tau_0}}}\leq \frac{2\kappa_3}{\sqrt{\kappa_5}}\bar\varepsilon_s,
\label{eq:proof-dual-opt-2}
\end{equation}
where $\kappa_3$ is the constant defined in the \textsc{GradEst} sub-routine. Subsequently, 
\begin{align}
\varepsilon_s &:= \|e_s\|_2=\|\hat\nabla Q(\lambda_s)-\nabla Q(\lambda_s)\|_2 = \big|\langle\lambda_s, A\hat D(\hat p(\lambda_s))-A D(p_\lambda^*)\rangle\big|\nonumber\\
&\leq \bar\lambda B_A(\|\hat D(\hat p(\lambda_s))-D(\hat p(\lambda_s))\|_2 + B_D\|\hat p(\lambda_s)-p_{\lambda_s}^*\|_2)\nonumber\\
&\leq \bar\lambda B_A\left(\frac{2\kappa_3}{\sqrt{\kappa_5}}\bar\varepsilon_s + \frac{\kappa_1 B_D}{\sqrt[4]{\kappa_5}}\sqrt{\bar\varepsilon_s}\right)\nonumber\\
&\leq  \bar\lambda B_A\left(\frac{2\kappa_3\kappa_6}{\sqrt{\kappa_5}}\times (1+\mu\eta_2)^{-s/2}+ \frac{\kappa_1 B_D\sqrt{\kappa_6}}{\sqrt[4]{\kappa_5}}\times (1+\mu\eta_2)^{-s/4}\right).\label{eq:proof-dual-opt-3}
\end{align}
Note that, by inductive hypothesis, Eq.~(\ref{eq:proof-dual-opt-3}) holds with $s$ replaced by $j$ for all prior iterations $j\leq s$ as well.

Invoke Lemma \ref{lem:inexact-md} and incorporate the upper bounds on $\|e_j\|_2$ for $j\leq s$ in Eq.~(\ref{eq:proof-dual-opt-3}). We have for iteration $s+1$ that
\begin{align}
\frac{\sum_{j=1}^{s+1} c_{j-1}[Q(\lambda_j)-Q(\lambda^*)]}{\sum_{j=1}^{s+1}c_{j-1}}& \leq  \frac{2(1+\mu\eta_2)^{-s}\bar\lambda^2}{\eta_2} + 2(1+\mu\eta_2)^{-s}\left(\sum_{j=0}^{s}\sqrt{\eta_2}(1+\mu\eta_2)^{j/2}\varepsilon_j\right)^2,
\label{eq:proof-dual-opt-4}
\end{align}
where $c_j=(1+\mu\eta_2)^j$.
For notational simplicity, denote 
$\delta_0 = B_\phi+\bar\lambda(B_A\bar d+\gamma_{\max})\sqrt{N}$, 
$\kappa_7=\bar\lambda B_A\times 2\kappa_3\kappa_6/\sqrt{\kappa_5}$ and $\kappa_8=\bar\lambda B_A\times \kappa_1 B_D\sqrt{\kappa_6}/\sqrt[4]{\kappa_5}$, so that $\|e_j\|_2\leq \kappa_7(1+\mu\eta_2)^{-j/2}+\kappa_8(1+\mu\eta_2)^{-j/4}$ for all $j\leq s$. Incorporating Eq.~(\ref{eq:proof-dual-opt-3}), Eq.~(\ref{eq:proof-dual-opt-4}) is simplified to
\begin{align}
\frac{\sum_{j=1}^{s+1} c_{j-1}[Q(\lambda_j)-Q(\lambda^*)]}{\sum_{j=1}^{s+1}c_{j-1}} &\leq\frac{2(1+\mu\eta_2)^{-s}\bar\lambda^2}{\eta_2} + 2(1+\mu\eta_2)^{-s}\left(\sum_{j=0}^{s}\sqrt{\eta_2}(\kappa_7+\kappa_8(1+\mu\eta_2)^{j/4}\right)^2\nonumber\\
&\leq \frac{2(1+\mu\eta_2)^{-s}\bar\lambda^2}{\eta_2} + 2\eta_2(1+\mu\eta_2)^{-s}\left(\kappa_7 s + \frac{\kappa_8(1+\mu\eta_2)^{(s+1)/4}}{\mu\eta_2}\right)^2\nonumber\\
&\leq \frac{4(\eta_2^{-1}\bar\lambda^2+\eta_2\kappa_7^2s^2)}{(1+\mu\eta_2)^s} + \frac{4\kappa_8^2\sqrt{1+\mu\eta_2}}{\mu^2\eta_2(1+\mu\eta_2)^{s/2}}\nonumber\\
&\leq \frac{8\kappa_7^2\ln^2 T}{\mu^2\eta_2(1+\mu\eta_2)^s} + \frac{4\kappa_8^2\sqrt{1+\mu\eta_2}}{\mu^2\eta_2(1+\mu\eta_2)^{s/2}}.\label{eq:proof-dual-opt-5}
\end{align}
Here, Eq.~(\ref{eq:proof-dual-opt-5}) holds because $s\leq \ln(T)/\ln(1+\mu\eta_2)\leq 2\ln(T)/(\mu\eta_2)$ almost surely.
With the choice that \footnote{Note the choice of $\kappa_3$ earlier and $\kappa_6$ later, $\kappa_5\geq 1$ holds.}
$$
\kappa_5 = \frac{32\kappa_3^2\kappa_6^2\bar\lambda B_A\ln^2 T}{\mu^2\eta_2\bar d\sqrt{N}}\vee \frac{16\kappa_1^4\kappa_6^2\bar\lambda^2B_D^4(1+\mu\eta_2)}{\mu^4\eta_2^2\bar d^2N},
$$
both terms on the right-hand side in Eq.~(\ref{eq:proof-dual-opt-5}) can be upper bounded by ${\delta_0}$ separately. Subsequently, 
\begin{equation}
\frac{\sum_{j=1}^{s+1} c_{j-1}[Q(\lambda_j)-Q(\lambda^*)]}{\sum_{j=1}^{s+1}c_{j-1}} \leq -(1+\mu\eta_2)^{s/2}\times 2\delta_0.
\label{eq:proof-dual-opt-5half}
\end{equation}
Note that
$$
\sum_{j=1}^{s+1}c_{j-1} = \sum_{j=0}^s (1+\mu\eta_2)^j\leq \frac{(1+\mu\eta_2)^{s+1}}{\mu\eta_2} \leq \frac{1+\mu\eta_2}{\mu\eta_2} c_s.
$$
Note also that $Q(\lambda_j)\geq Q(\lambda^*)$ for all $j$ thanks to the optimality of $\lambda^*$. Subsequently, Eq.~(\ref{eq:proof-dual-opt-5half}) is reduced to
\begin{align}
Q(\lambda_{s+1})-Q(\lambda^*) &\leq \frac{(1+\mu\eta_2)^{3/2}}{\mu\eta_2}\times 2\delta_0(1+\mu\eta_2)^{-(s+1)/2}.
\end{align}
With the choice that
$$
\kappa_6 \geq \frac{2\delta_0(1+\mu\eta_2)^{3/2}}{\mu\eta_2} = \frac{2(B_\phi+\bar\lambda(B_A\bar d+\gamma_{\max})\sqrt{N})(1+\mu\eta_2)^{3/2}}{\mu\eta_2},
$$
we have that $Q(\lambda_{s+1})-Q(\lambda)^*\leq \kappa_6(1+\mu\eta_2)^{-(s+1)/2}$, 
which is to be proved.

In the remainder of this section we present the proof of Lemma \ref{lem:inexact-md}.
\begin{proof}{Proof of Lemma \ref{lem:inexact-md}.}
Let $L=1/\eta_2=B_A^2/\sigma_\phi$. By Lemma \ref{lem:strong-convexity}, $Q$ is $L$-strongly smooth and $\mu$-strongly convex, meaning that $\|\nabla Q(\lambda)-\nabla Q(\lambda')\|_2\leq L\|\lambda-\lambda\|_2$
and $Q(\lambda')\geq Q(\lambda) + \langle\nabla Q(\lambda),\lambda'-\lambda\rangle + \frac{\mu}{2}\|\lambda'-\lambda\|_2^2$ for any $\lambda,\lambda'\in\Lambda$.
Let also $\zeta=1/(\mu\eta_2) = L/\mu$ be the condition number of $Q$.

Note that $e_s = \hat\nabla Q(\lambda_s)-\nabla Q(\lambda_s) = \hat\nabla h(\lambda_s)-\nabla h(\lambda_s)$, where $h(\lambda_s) = Q(\lambda_s) - \frac{\mu}{2}\|\lambda_s\|_2^2$.
Let $\phi_s(\lambda) := \langle\hat\nabla h(\lambda_s),\lambda\rangle + \frac{\mu}{2}\|\lambda\|_2^2 + \frac{1}{2\eta_2}\|\lambda-\lambda_s\|_2^2$.
It is easy to verify that $\phi_s(\cdot)$ is continuously differentiable and $(\mu+\eta_2^{-1})$-strongly convex.
By Line \ref{line:pgd-dual} in Algorithm \ref{alg:dual-opt}, we know that $\lambda_{s+1}=\arg\min_{\lambda\in\Lambda}\phi_s(\lambda)$.
The first-order KKT condition yields that $\langle\nabla\phi_s(\lambda_{s+1}),\lambda-\lambda_{s+1}\rangle\geq 0$ for all $\lambda\in\Lambda$,
and furthermore by the strong convexity of $\phi_s$ it holds that $\phi_s(\lambda)-\phi_s(\lambda_{s+1})\geq \frac{\mu+\eta_2^{-1}}{2}\|\lambda-\lambda_s\|_2^2$ for all $\lambda\in\Lambda$.
Subsequently,
\begin{align}
\langle\hat\nabla h(\lambda_s), &\lambda-\lambda_{s+1}\rangle + \frac{\mu}{2}\|\lambda\|_2^2 - \frac{\mu}{2}\|\lambda_{s+1}\|_2^2 + \frac{1}{2\eta_2}\left(\|\lambda-\lambda_s\|_2^2-\|\lambda_{s+1}-\lambda_s\|_2^2\right)\geq \left(\frac{1}{2\eta_2}+\frac{\mu}{2}\right)\|\lambda-\lambda_{s+1}\|_2^2.\label{eq:proof-inexact-md-1}
\end{align}
On the other hand, because $h$ is continuously differentiable with $L$-Lipschitz continuous gradients, it holds that
\begin{align}
h(\lambda_{s+1}) \leq h(\lambda_s) + \langle\nabla h(\lambda_s),\lambda_{s+1}-\lambda_s\rangle + \frac{L}{2}\|\lambda_{s+1}-\lambda_s\|_2^2.\label{eq:proof-inexact-md-2}
\end{align}
Additionally, using the definition of $e_s$, we have for every $\lambda\in\Lambda$ that
\begin{align}
\langle\hat\nabla h(\lambda_s),\lambda-\lambda_{s+1}\rangle& + \langle\nabla h(\lambda_s),\lambda_{s+1}-\lambda_s\rangle = \langle\nabla h(\lambda_s),\lambda-\lambda_s\rangle + \langle e_s,\lambda-\lambda_{s+1}\rangle\nonumber\\
&\leq h(\lambda)-h(\lambda_s) + \|e_s\|_2\|\lambda-\lambda_{s+1}\|_2,\label{eq:proof-inexact-md-3}
\end{align}
where in Eq.~(\ref{eq:proof-inexact-md-3}) we use the convexity of $h(\cdot)$ and the Cauchy-Schwarz inequality.
Combining Eqs.~(\ref{eq:proof-inexact-md-1},\ref{eq:proof-inexact-md-2},\ref{eq:proof-inexact-md-3}) and using the fact that $\eta_2=1/L$, we obtain
\begin{align}
\frac{\mu+\eta_2^{-1}}{2}\|\lambda-\lambda_{s+1}\|_2^2 + Q(\lambda_{s+1}) &\leq Q(\lambda) + \frac{L-\eta_2^{-1}}{2}\|\lambda_{s+1}-\lambda_s\|_2^2 + \frac{1}{2\eta_2}\|\lambda-\lambda_s\|_2^2 + \|e_s\|_2\|\lambda-\lambda_{s+1}\|_2\nonumber\\
&\leq  Q(\lambda) +  \frac{1}{2\eta_2}\|\lambda-\lambda_s\|_2^2 + \|e_s\|_2\|\lambda-\lambda_{s+1}\|_2.\label{eq:proof-inexact-md-4}
\end{align}

Next, recall the definition of the averaging coefficients that $c_s=(1+\mu\eta_2)^s$, so that $c_0=1$ and $c_s(\mu+\eta_2^{-1})=c_{s+1}\eta_2^{-1}$. Invoking Eq.~(\ref{eq:proof-inexact-md-4})
and setting $\lambda=\lambda^*=\arg\min_{\lambda\in\Lambda}Q(\lambda)$, we have
\begin{align}
\frac{c_{s+1}}{2\eta_2}\|\lambda_{s+1}-\lambda^*\|_2^2 + c_s(Q(\lambda_{s+1})-Q(\lambda^*)) \leq \frac{c_s}{2\eta_2}\|\lambda_s-\lambda^*\|_2^2 + c_s\|e_s\|_2\|\lambda_{s+1}-\lambda^*\|_2.
\label{eq:proof-inexact-md-5}
\end{align}
Summing both sides of Eq.~(\ref{eq:proof-inexact-md-5}) over $s=0,1,\cdots,m-1$ and telescoping, we obtain
\begin{align}
\frac{c_m}{\eta_2}\|\lambda_m-\lambda^*\|_2^2 \leq \frac{c_0}{2\eta_2}\|\lambda_0-\lambda^*\|_2^2 + \sum_{s=1}^m c_{s-1}\|e_{s-1}\|_2\|\lambda_s-\lambda^*\|_2.\label{eq:proof-inexact-md-6}
\end{align}

Now, let $v_s := \sqrt{c_s\eta_2^{-1}\|\lambda_s-\lambda^*\|_2^2/2}$, $a := \eta_2^{-1}\|\lambda_0-\lambda^*\|_2^2/2$ and $b_s := c_{s-1}\|e_{s-1}\|_2/\sqrt{c_{s-1}\eta_2^{-1}}$ for $s=1,2,\cdots,m$.
From Eq.~(\ref{eq:proof-inexact-md-6}), it is easy to verify that $v_t^2\leq a+\sum_{s=1}^t b_sv_s$ for all $t\geq 1$, and $v_0^2\leq a$.
Let $A_0 := a$ and $A_t := a + \sum_{s=1}^t b_sv_s$ for all $t\geq 1$, so that $v_t\leq \sqrt{A_t}$ and $\{A_t\}_{t\geq 0}$ is non-negative and non-decreasing. Then, for all $t\geq 0$, 
$$
A_{t+1}=A_t + b_{t+1}v_{t+1}\leq A_t + b_{t+1}\sqrt{A_{t+1}},
$$
and therefore
$$
\sqrt{A_{t+1}}\leq A_t/\sqrt{A_{t+1}} + b_{t+1} \leq \sqrt{A_t}+b_{t+1}.
$$
Summing both sides of the above inequality over $s=0,1,\cdots,m-1$, we obtain
$$
\sqrt{A_m}\leq \sqrt{A_0}+\sum_{s=1}^m b_s = \sqrt{a} + \sum_{s=1}^m b_s.
$$
Subsequently,
\begin{align}
\sum_{s=1}^m c_{s-1}(Q(\lambda_s)-Q(\lambda^*)) &\leq \left(\sqrt{\frac{\|\lambda_0-\lambda^*\|_2^2}{2\eta_2}} + \sum_{s=1}^m\frac{\sqrt{\eta_2}c_{s-1}\|e_{s-1}\|_2}{\sqrt{c_s}}\right)^2\nonumber\\
&\leq \frac{1}{\eta_2}\|\lambda_0-\lambda^*\|_2^2 + 2\left(\sum_{s=1}^m \sqrt{\eta_2 c_{s-1}}\|e_{s-1}\|_2\right)^2.
\label{eq:proof-inexact-md-7}
\end{align}

Finally, note that
\begin{align}
\frac{1}{\sum_{s=0}^{m-1}c_s} &= \frac{\mu\eta_2}{(1+\mu\eta_2)^m-1} = \frac{\mu\eta_2}{(1+\mu\eta_2)^m}\frac{1}{1-(1+\mu\eta_2)^{-m}} \leq (1+\mu\eta_2)^{-m+1}.\label{eq:proof-inexact-md-8}
\end{align}
Combining Eqs.~(\ref{eq:proof-inexact-md-6},\ref{eq:proof-inexact-md-7}) we complete the proof of Lemma \ref{lem:inexact-md}.
\end{proof}

\section{Proof of results in Section~\ref{sec:regret}}\label{sec:proof_main_thm}

\subsection{Proof of Theorem \ref{thm:regret-upper-bound}}
Note that to prove Theorem \ref{thm:regret-upper-bound}, it suffices to upper bound the cumulative deviation from the optimal price $f(p^*)-f(p_t)$, as well as the cumulative ``over-demand''
so that the inventory of all resources will not be depleted too early.
Fix an iteration $s$ in \textsc{DualOpt} and an iteration $\tau$ in \textsc{PrimalOpt}, and let $\mT(s,\tau)$ be the set of time periods, which satisfies $|\mT(s,\tau)|=n_\tau=\lceil (1-\tilde\eta)^{-2\tau} n_0\rceil$.
Let also $p$ be the input of \textsc{GradEst}.
For the first half of $\mathcal T(s,\tau)$ (denoted as $\mT_1(s,\tau)$, we have for some $p'\in[\underline p,\overline p]$ that
\begin{align}
\sum_{t\in\mT_1(s,\tau)} f(p_t) &= \frac{n_\tau}{4N}\sum_{i=1}^N f(p+ue_i) + f(p-ue_i)\geq \frac{n_\tau}{4N}\sum_{i=1}^N 2f(p) - u^2 \big|e_i^\top \nabla^2 f(p')e_i\big|\label{eq:proof-theorem-1}\\
&\geq \frac{n_\tau}{2} f(p) - \frac{n_\tau}{4} u^2 \|\nabla^2 f(p')\|_\op\geq \frac{n_\tau}{2} f(p) - \frac{n_\tau}{4}B_fu^2\label{eq:proof-theorem-2}\\
&\geq \frac{n_\tau}{2}f(p) - \frac{n_\tau}{4}\times \frac{B_f N}{\sqrt{n_\tau}} =\frac{n_\tau}{2}f(p) - \frac{B_f N}{4\sqrt{n_\tau}}. \label{eq:proof-theorem-3}
\end{align}
Here, the second inequality of Eq.~(\ref{eq:proof-theorem-1}) holds by the Taylor expansion of $f$ at $p$ with Lagrange remainders;
the second inequality of Eq.~(\ref{eq:proof-theorem-2}) holds because $\|\nabla^2 f(p')\|_\op\leq B_f$ for all $p'$, thanks to Assumption \ref{asmp:f-phi};
the first inequality in Eq.~(\ref{eq:proof-theorem-3}) holds because $u=\sqrt{N}/\sqrt[4]{n}$, thanks to the third property of Lemma \ref{lem:primal-opt}.

Similarly, the expected resource consumptions over the first half of $\mT(s,\tau)$ can be analyzed as
\begin{align}
\sum_{t\in\mT_1(s,\tau)} \langle a_j,D(p_t)\rangle &= \frac{n_\tau}{4N}\sum_{i=1}^N \langle a_j,D(p+ue_i) + D(p-ue_i)\rangle\nonumber\\
&  = \frac{n_\tau}{4N}\sum_{i=1}^N 2\langle a_j, D(p)\rangle \pm 2L_Du^2\|a_j\|_2\nonumber\\
& = \frac{n_\tau}{2}\langle a_j, D(p)\rangle \pm \frac{L_D N \|a_j\|_2}{2\sqrt{n_\tau}},\;\;\;\;\;\;\forall j\in[m].\label{eq:proof-theorem-4}
\end{align}

Next, let $\mT_2(s,\tau)=\mT(s,\tau)\backslash\mT_1(s,\tau)$ be the second half of $\mT(s,\tau)$, where the price vectors are committed to $p_\tau\equiv \tilde p$ for all $\tau\in\mT_2(s,\tau)$,
with $\tilde p$ being obtained on Line \ref{line:tilde-p} in the \textsc{GradEst} sub-routine. By Lemma \ref{lem:primal-opt}, it holds with probability $1-\tilde O(T^{-1})$ that
\begin{align}
\frac{f(p)+f(\tilde p)}{2} &\geq f(p^*) - \frac{1.2\kappa_1^2 (B_f+B_AB_J\bar\lambda) N + (2\kappa_3+\sqrt{\kappa_5})\bar\lambda\sqrt{N}}{\sqrt{n_\tau}};\label{eq:proof-theorem-5}\\
\frac{\langle a_j, D(p)+D(\tilde p)\rangle}{2} &\leq \gamma_j + \frac{\kappa_3}{\sqrt{n_\tau}}, \;\;\;\;\;\forall j\in[M].\label{eq:proof-theorem-6}
\end{align}
Combine Eqs.~(\ref{eq:proof-theorem-3},\ref{eq:proof-theorem-4},\ref{eq:proof-theorem-5},\ref{eq:proof-theorem-6}). We obtain
\begin{align}
\sum_{t\in\mT(s,\tau)}f(p_t) &\geq n_\tau f(p^*) - \kappa_8\sqrt{n_\tau};\label{eq:proof-theorem-7}\\
\sum_{t\in\mT(s,\tau)}\langle a_j, D(p_t)\rangle &\leq n_\tau\gamma_j + (\kappa_3+L_DN\|a_j\|_2)\sqrt{n_\tau},\;\;\;\;\;\;\forall j\in[M],\label{eq:proof-theorem-8}
\end{align}
with 
\begin{align*}
\kappa_8 &:= B_f N + 1.2\kappa_1^2 (B_f+B_AB_J\bar\lambda) N + (2\kappa_3+\sqrt{\kappa_5})\bar\lambda\sqrt{N} \\
&= O(N^{3.25}\ln^{1.5}(NT)+N^{2.5}\ln^3(NT)).
\end{align*}

Let $s_0$ be the last iteration in \textsc{DualOpt}, and $\tau_0(s)$ be the last iteration in \textsc{PrimalOpt} when the outer iteration in \textsc{DualOpt} is $s\leq s_0$.
It is easy to verify that $s_0\leq \ln(T)/\ln(1+\eta\mu_2)\leq 2\ln(T)/(\eta\mu_2)$ and $\tau_0(s)\leq \tau_0 = \ln(T)/(-\ln(1-\tilde\eta))\leq 2\ln(T)/\tilde\eta$ almost surely for all $s$, where $\tilde\eta=\eta_1\sigma_D^2\sigma_\phi/2$.
Note also that $\sum_{s,\tau}|\mT(s,\tau)|\leq T$. Subsequently, applying Cauchy-Schwarz inequality, Eqs.~(\ref{eq:proof-theorem-7},\ref{eq:proof-theorem-8}) lead to
\begin{align}
\sum_{t=1}^T f(p^*)-f(p_t) &\leq \kappa_8\sqrt{\frac{4\ln^2 T}{\mu\eta_2\tilde\eta}}\times \sqrt{T}=:\kappa_9\sqrt{T};\label{eq:proof-theorem-9}\\
\sum_{t=1}^T\langle a_j, D(p_t)\rangle &\leq \gamma_j T + (\kappa_3+L_D N\|a_j\|_2)\sqrt{\frac{4\ln^2 T}{\mu\eta_2\tilde\eta}}\times \sqrt{T}\nonumber\\
& =: \gamma_j T+\kappa_{10}(j)\sqrt{T},\;\;\;\;\;\;\forall j\in[M].\label{eq:proof-theorem-10}
\end{align}
where 
\begin{align*}
\kappa_9 &:= \frac{2\kappa_8 \ln(T)}{\sqrt{\mu\eta_2\tilde\eta}}\;\;\;\;\;\;\text{and}\;\;\;\;\;\kappa_{10}(j) := \frac{2(\kappa_3+L_D N\|a_j\|_2)\ln(T)}{\sqrt{\mu\eta_2\tilde\eta}}.
\end{align*}

Note that Eq.~(\ref{eq:proof-theorem-10}) together with $\gamma_j\geq\gamma_{\min}>0$ and the Hoeffding-Azuma's inequality, implies that inventory will not be depleted until the last $2\kappa_{10}(j)\sqrt{T}/\gamma_{\min}$ time periods, during which an $B_r$ amount of revenue per period might be lost due to insufficient resources. Subsequently, the total regret can be upper bounded by
$$
\kappa_9\sqrt{T} + \frac{2B_r\max_j\kappa_{10}(j)}{\gamma_{\min}}\sqrt{T},
$$
which is to be proved.

\end{document}